\documentclass[]{style}

\usepackage[toc,page,header]{appendix}
\usepackage{enumitem}
\usepackage{xcolor}
\usepackage[dvipsnames]{xcolor}
\usepackage{listings}
\usepackage{caption}

\usepackage{natbib}

\usepackage{CJKutf8}

\usepackage{xargs}  

\usepackage{todonotes}  

\usepackage{multirow}

\usepackage{cleveref}

\usepackage{amsmath}
\usepackage{dsfont}

\usepackage{svg}

\usepackage{mathrsfs}
\usepackage{adjustbox}
\usepackage{multirow}
\usepackage{multicol}
\usepackage{tcolorbox}
\usepackage{changepage}
\usepackage{enumitem}
\usepackage{graphicx}
\usepackage{amssymb}
\usepackage{xcolor}
\usepackage{float}
\usepackage{multirow}
\usepackage{threeparttable}
\usepackage{graphicx}
\usepackage{subcaption}
\usepackage{algorithm}
\usepackage{algpseudocode}
\usepackage{wrapfig}
\usepackage[table]{xcolor}  
\usepackage{colortbl}       
\usepackage{tabularx} 
\usepackage{makecell} 
\usepackage[dvipsnames]{xcolor}

\newcolumntype{g}{>{\columncolor{gray!10}}c} 
\newcolumntype{A}{>{\raggedright\arraybackslash}p{4.5cm}} 
\newcolumntype{Y}{>{\centering\arraybackslash}X} 

\definecolor{catgray}{gray}{0.9}
\definecolor{skyblue}{rgb}{0.53,0.81,0.92} 

\colorlet{skyblue!30}{skyblue!30!white} 

\definecolor{customblue}{RGB}{70,130,180}  

\newtcolorbox{evolbox}[2][]{%
  enhanced,
  colframe=customblue,
  colback=white,
  coltitle=white,
  rounded corners,
  boxrule=1pt,
  titlerule=0pt,
  toptitle=1mm,
  bottomtitle=1mm,
  fonttitle=\bfseries,
  width=#2\textwidth, 
  #1
}

\usepackage{minitoc}
\usepackage{url}

\PassOptionsToPackage{table,xcdraw}{xcolor}
\usepackage{titletoc}
\usepackage{placeins}
\usepackage{pifont}

\definecolor{RowBlue}{HTML}{E9F2FB}
\definecolor{RowRed}{HTML}{F9EAEA}
\definecolor{Top1}{HTML}{50DB4B} 
\definecolor{Top2}{HTML}{A5FFA2} 
\definecolor{Top3}{HTML}{D9FFD9} 
\definecolor{Sub1}{HTML}{EAB8B8}
\definecolor{Sub2}{HTML}{E4E4E4}
\newcommand{\cmark}{\textcolor{green!60!black}{\checkmark}}
\newcommand{\xmark}{\textcolor{red!70!black}{\ding{55}}}

\newcommand{\myparagraph}[1]{\textbf{#1}\hspace{1.8ex}}
\newcommand{\abpar}[2]{\textbf{\colorbox{#1}{#2}}}
\definecolor{gearred}{HTML}{D85140}
\definecolor{reprablue}{HTML}{5384ED}
\definecolor{steorange}{HTML}{EF8444}
\definecolor{softgreen}{HTML}{658E40}

\renewcommand{\emph}[1]{\textit{#1}}

\definecolor{codepink}{RGB}{220,20,120}
\definecolor{codegreen}{RGB}{0,150,0}
\definecolor{codegray}{RGB}{140,140,140}
\definecolor{codeorange}{RGB}{230,120,60}

\lstdefinestyle{pytorchstyle}{
    language=Python,
    basicstyle=\ttfamily\small,
    keywordstyle=\color{codepink}\bfseries,
    commentstyle=\color{codegray}\itshape,
    stringstyle=\color{codeorange},
    numberstyle=\tiny\color{codegray},
    numbers=none,
    showstringspaces=false,
    breaklines=true,
    frame=none,
    columns=fullflexible,
    keepspaces=true,
    xleftmargin=1.5em
}

\newcommand{\boldtitle}[1]{{\bfseries #1}}
\title{\boldtitle{G}uided \boldtitle{E}nd-to-End \boldtitle{A}uto\boldtitle{R}egression for Image Synthesis}

\author{
  Bin Lin\texorpdfstring{$^{1,2,^*}$}{}
  \hspace{0.1cm}
  Zheyuan Liu\texorpdfstring{$^{1,2,^*}$}{}
  \hspace{0.1cm}
  Chenguo Lin\texorpdfstring{$^{1}$}{}
  \hspace{0.1cm}
  Sixiang Chen\texorpdfstring{$^{2,^*}$}{}
  \hspace{0.1cm}
  Yunyang Ge\texorpdfstring{$^{1,2,^*}$}{}
  \hspace{0.1cm} \\
  Yunlong Lin\texorpdfstring{}{}
  \hspace{0.1cm}
  Jianwei Zhang\texorpdfstring{$^{2}$}{} \hspace{0.1cm}
  Miles Yang{$^{2}$}{}
  \hspace{0.1cm}
  Zhao Zhong{$^{2}$}{}
  \hspace{0.1cm}
  Liefeng Bo{$^{2}$}{}
  \hspace{0.1cm}
  Li Yuan\texorpdfstring{$^{1,\dag}$}{}
  \hspace{0.1cm}
}

\makeatletter
\g@addto@macro\authorlist{\\[2mm]
  {\small
  \texorpdfstring{$^{*}$}{}Work done during internship at Tencent Hunyuan \quad
  \texorpdfstring{$^{\dag}$}{}Corresponding author
  }\\[0.2mm]
}
\makeatother

\affiliation[1]{Peking University}
\affiliation[2]{Tencent Hunyuan}

\newcommand{\answerTODO}[1][]{\textcolor{red}{\bfseries [TODO]}}
\newcommand{\justificationTODO}[1][]{\textcolor{red}{\bfseries [TODO]}}

\abstract{
Visual generative models are typically trained in two stages. A tokenizer is first trained for reconstruction and then frozen, after which a generator is trained on its discrete indices or continuous latents. This decoupling leaves the tokenizer unaware of what the generator finds easy to model. We present \textbf{GEAR} (\textbf{G}uided \textbf{E}nd-to-end \textbf{A}uto\textbf{R}egression), which trains a vector-quantized (VQ) tokenizer and an autoregressive (AR) generator jointly and end-to-end, guided by representation alignment. The key obstacle is that the VQ index fed to the AR model is non-differentiable, so gradients cannot reach the tokenizer, and a straight-through estimator collapses. GEAR resolves this with a dual read-out of the codebook assignment. A hard, one-hot branch trains the AR with next-token prediction, while a differentiable soft branch carries a representation-alignment loss that flows back to guide only the tokenizer. The AR model thereby steers its tokenizer toward an index distribution it can predict more easily. This shifts the alignment burden from the tokenizer to the AR: the tokenizer's own features become \emph{less} DINOv2-like while the AR's become more so, the opposite of diffusion-side recipes that make the latent itself semantic. GEAR speeds up ImageNet gFID convergence by up to $10\times$ relative to the strong LlamaGen-REPA baseline, learns markedly better patch-level and spatially-coherent features, and generalizes across quantizers (VQVAE, LFQ, IBQ) and to text-to-image generation.
}

\checkdata[
\raisebox{-0.3em}{\includegraphics[width=0.025\linewidth]{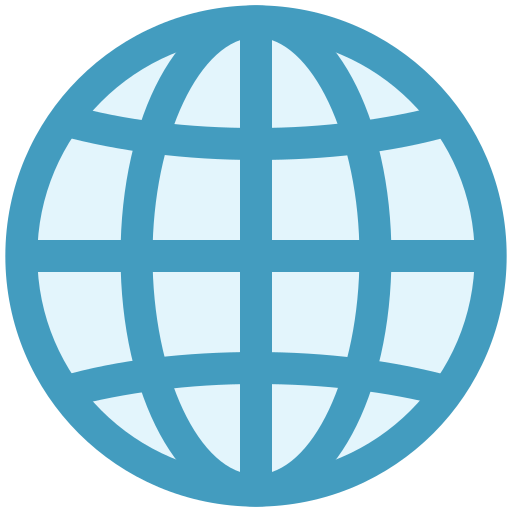}}~~Project Page]{\href{https://linb203.github.io/gear}{\texttt{https://linb203.github.io/gear}}
\\[-2.3ex]}

\checkdata[
\raisebox{-0.3em}{\includegraphics[width=0.025\linewidth]{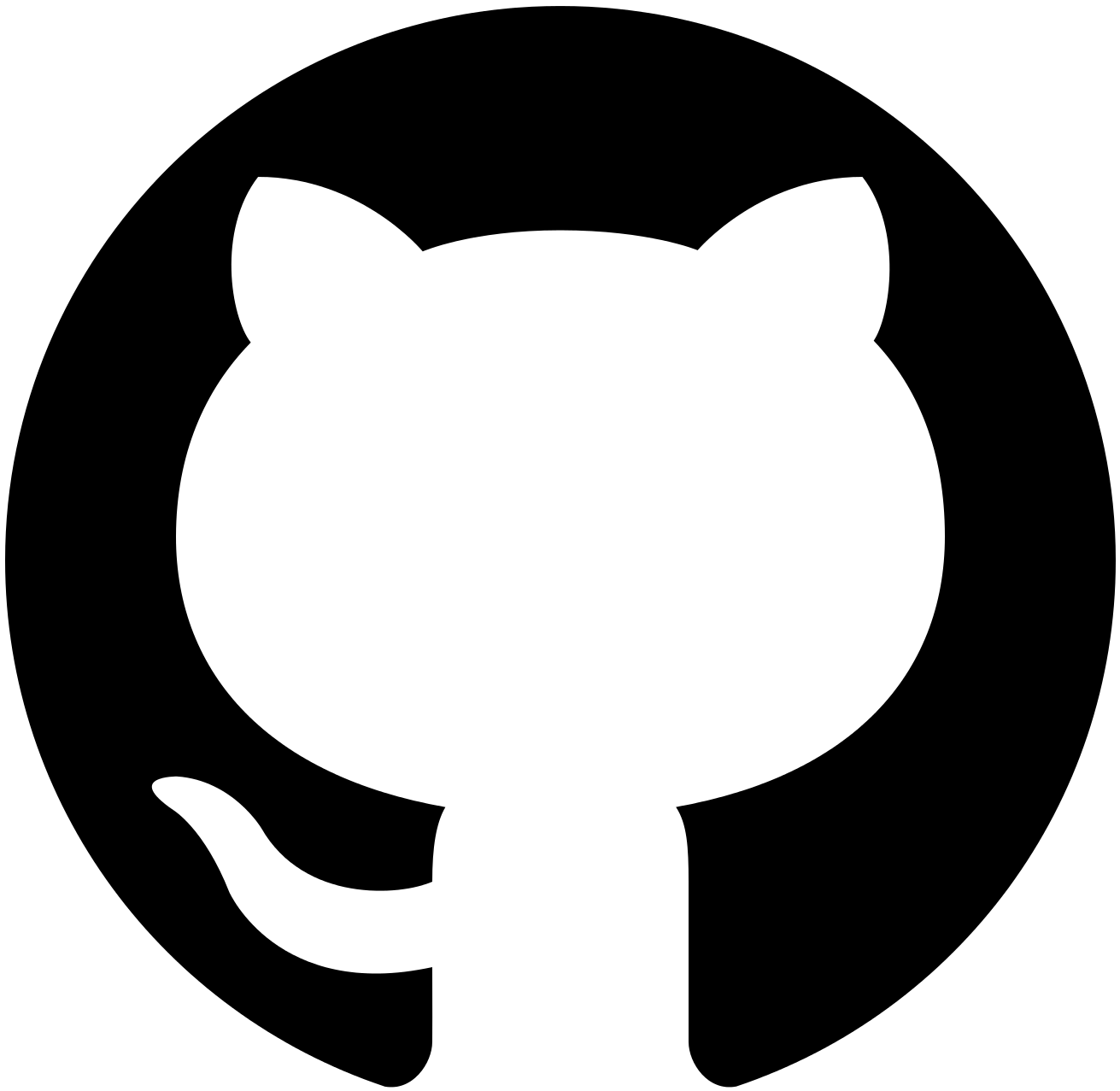}}~~GitHub Repo]{\href{https://github.com/Tencent-Hunyuan/GEAR}{\texttt{https://github.com/Tencent-Hunyuan/GEAR}}
\\[-2.3ex]}

\checkdata[
\raisebox{-0.3em}{\includegraphics[width=0.025\linewidth]{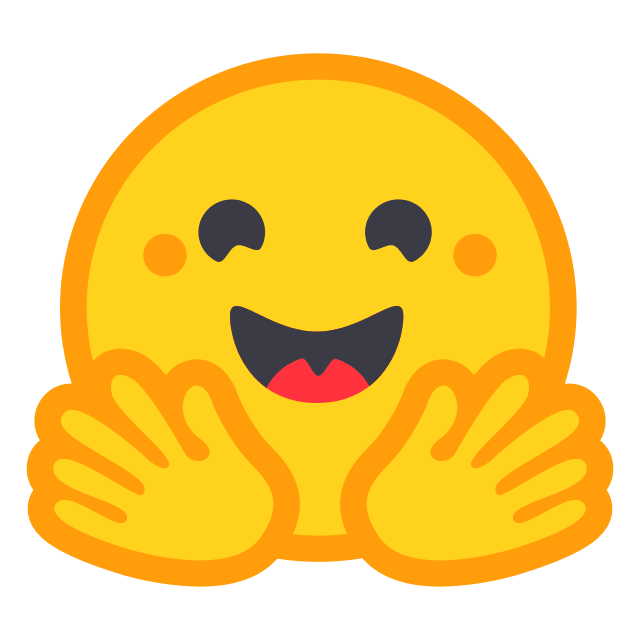}}~~HuggingFace Model]{\href{https://huggingface.co/collections/BinLin203}{\texttt{https://huggingface.co/collections/BinLin203}}
\\[-2.3ex]}

\begin{document}
\maketitle

\vspace{-7mm}
\noindent\begin{minipage}{\linewidth}
\centering
\includegraphics[width=1.0\linewidth]{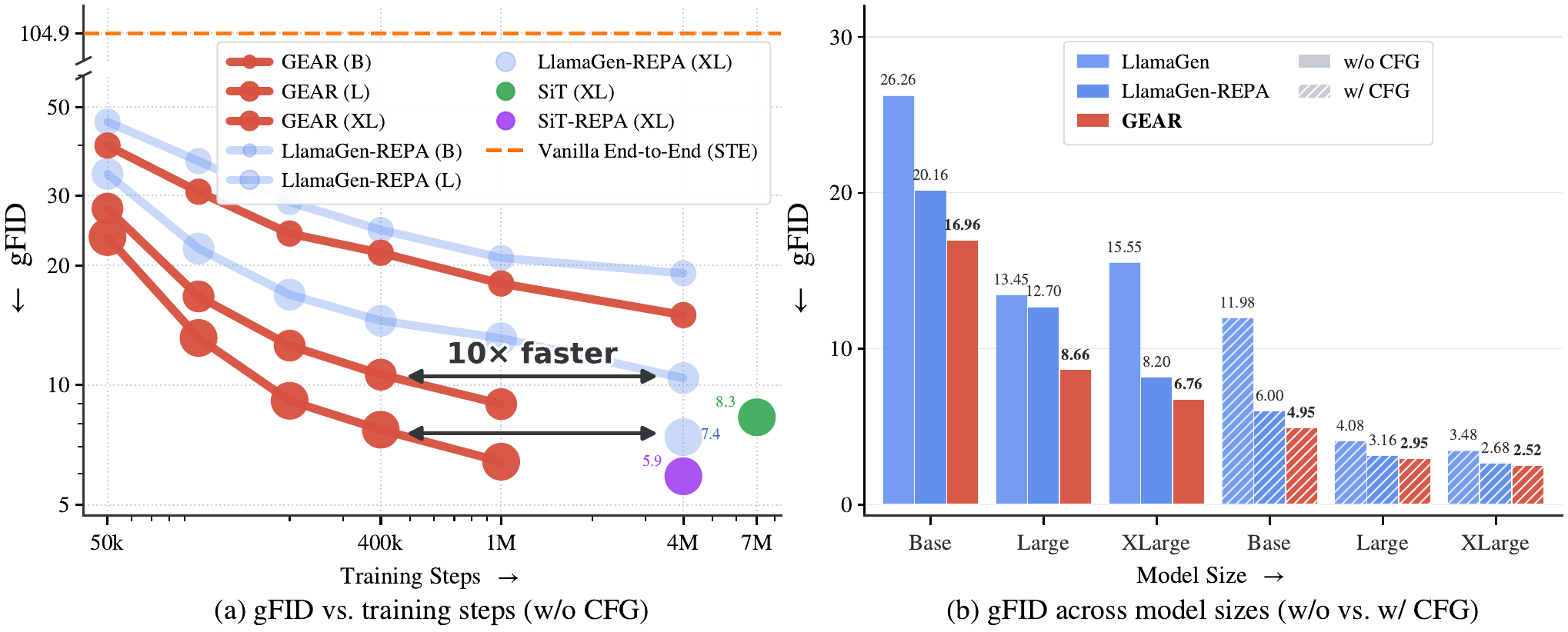}
\captionof{figure}{\textbf{GEAR accelerates and improves autoregressive image generation.} \textbf{(a)} gFID versus training steps on ImageNet (without CFG): \textcolor{gearred}{GEAR} converges up to $10\times$ faster than \textcolor{reprablue}{LlamaGen-REPA}, whereas the naive end-to-end variant that back-propagates into the tokenizer through the \textcolor{steorange}{straight-through estimator} diverges (gFID${\approx}105$). \textbf{(b)} gFID across model scales at $1.5$M steps: \textcolor{gearred}{GEAR} improves performance at every size (B/L/XL), both without and with CFG.}
\label{fig:teaser}
\end{minipage}

\section{Introduction}
\label{sec:intro}

Modern visual generative models are almost universally trained in two stages. This holds both for autoregressive (AR) transformers over discrete tokens~\citep{llamagen,var} and for diffusion models over continuous latents~\citep{dit,sit}. A tokenizer, either a VQ-VAE~\citep{vqvae} or a continuous VAE, is first trained to reconstruct images. It is then \emph{frozen}, and a generator is trained on the resulting indices or latents. The tokenizer is therefore optimized purely for reconstruction, oblivious to whether the latent space it induces is easy for the downstream generator to model.

This separation is convenient but suboptimal. The latent distribution is fixed by a reconstruction objective that does not know whether the induced sequence is easy to generate. The two goals are also in tension. Faithful reconstruction favors high-variance, detail-rich latents, whereas generation favors simple, predictable structure. Recent work begins to dissolve this boundary in the diffusion setting. A line of \emph{representation alignment} methods accelerates training by injecting external semantics. REPA~\citep{repa} aligns a diffusion model's intermediate features with a pretrained encoder such as DINOv2~\citep{dinov2}, while VA-VAE and MAETok~\citep{vavae,maetok} align the VAE \emph{latent} itself with such features. Going further, REPA-E~\citep{repae} trains the VAE and the diffusion model jointly end-to-end. Crucially, it shows that naively back-propagating the diffusion loss into the VAE \emph{fails}, because the denoising objective flattens the latent variance that reconstruction needs. REPA-E therefore tunes the VAE end-to-end through the alignment loss rather than the diffusion loss.

The same opportunity exists for AR generation over discrete VQ tokens, but it is fundamentally harder. The map from a VQ index to the AR input is a non-differentiable $\arg\max$, so one cannot back-propagate any signal from the AR generator to the tokenizer, and the obvious remedy, a straight-through estimator (STE), is unstable and collapses the codebook in our joint setting (gFID${\approx}105$, \cref{fig:teaser}). We propose \textbf{GEAR} (\textbf{G}uided \textbf{E}nd-to-end \textbf{A}uto\textbf{R}egression), which trains the VQ tokenizer and the AR generator jointly and end-to-end and resolves this with a dual read-out of the per-position codebook assignment. A \emph{hard}, one-hot read-out reproduces the discrete tokens used at inference and trains the AR with next-token prediction and an alignment loss. A \emph{soft} read-out is a temperature-weighted interpolation over the nearest codewords, and is therefore differentiable, carrying an alignment loss that flows back to update \emph{only} the tokenizer. We never route the NTP loss into the tokenizer, because letting it reshape the codebook invites a collapse to a few low-entropy codes that trades reconstruction for predictability, as the concurrent EOSTok~\citep{eostok} also reports. The differentiable soft read-out is what makes any end-to-end gradient possible in this discrete setting, succeeding precisely where the STE collapses.

What this guidance actually does is, perhaps surprisingly, the opposite of the diffusion-side recipe. On the diffusion side, REPA-E, VA-VAE and MAETok make the \emph{latent} more semantic by aligning it to a pretrained encoder. In GEAR the tokenizer's own features instead become \emph{less} DINOv2-like, most strongly at the patch level (\cref{tab:tok_align}). Rather than turning semantic, the tokenizer re-organizes its discrete index distribution toward a more predictable, lower-entropy usage (\cref{fig:soft}), without sacrificing reconstruction. The semantic alignment instead emerges inside the \emph{AR generator}, whose hidden states track DINOv2 far more closely per patch and carry markedly more locally-coherent, spatially-causal structure, where LlamaGen-REPA~\citep{llamagen-repa} attains only global, image-level alignment (\cref{fig:repr_analysis}). End-to-end guidance thus shifts the alignment burden from the tokenizer to the AR: the tokenizer need not look semantic, only emit tokens the AR can predict, and this local, patch-level structure is exactly what makes next-token prediction easy, which explains GEAR's faster convergence and higher sample quality.

Our contributions are summarized as follows.
\begin{itemize}[leftmargin=1.2em,itemsep=2pt,topsep=2pt]
\item \textbf{Guided end-to-end training.} We introduce GEAR, which jointly trains a VQ tokenizer and an AR generator. A differentiable soft-assignment bridge lets the AR generator's representation-alignment objective guide the tokenizer. This overcomes the non-differentiable index that defeats the straight-through estimator, and speeds up ImageNet gFID convergence by up to $10\times$ relative to LlamaGen-REPA (\cref{fig:teaser}).
\item \textbf{Where representation alignment lives.} A representation analysis (\cref{sec:repr_analysis}) shows that end-to-end guidance shifts the alignment burden from the tokenizer to the AR. Unlike diffusion-side recipes, GEAR's tokenizer becomes \emph{less} DINOv2-like and re-organizes its index distribution toward predictable tokens, while the AR's features become more DINOv2-like \emph{per patch}, with reconstruction preserved.
\item \textbf{Generality.} The same mechanism works across quantizers, including VQVAE, LFQ~\citep{magvitv2,openmagvitv2}, and IBQ~\citep{ibq}. The end-to-end-tuned tokenizer is a drop-in that transfers across settings, and end-to-end training on ImageNet also accelerates text-to-image generation.
\end{itemize}

\section{Related Work}
\label{sec:related_work}

\noindent\textbf{Tokenizers and two-stage visual generation.}
Modern visual generators are built on a tokenizer that maps an image to a compact space, followed by a generative model over that space. Discrete pipelines pair a VQ tokenizer, such as VQ-VAE~\citep{vqvae}, VQGAN~\citep{vqgan}, LFQ~\citep{magvitv2,openmagvitv2}, or IBQ~\citep{ibq}, with an autoregressive model~\citep{llamagen,var} or a masked generator~\citep{maskgit}. Continuous pipelines pair a VAE with a diffusion transformer~\citep{dit,sit}. In nearly all of these, the tokenizer is trained for reconstruction and then frozen, so the generator inherits a latent space it cannot influence. A few methods sidestep the discrete bottleneck by generating continuous tokens autoregressively~\citep{mar}, but the dominant, LLM-style route keeps discrete VQ tokens. We follow this discrete route and instead ask how the tokenizer itself should be shaped for it.

\noindent\textbf{Representation alignment and semantic tokenizers.}
Semantics from self-supervised encoders such as DINOv2~\citep{dinov2}, DINOv3~\citep{dinov3}, SigLIPv2~\citep{siglip2}, and V-JEPA2.1~\citep{vjepa21} have become a powerful prior for generation. One family aligns the generator or the latent to such an encoder. REPA~\citep{repa} aligns a diffusion model's intermediate features to DINOv2, while VA-VAE~\citep{vavae} and MAETok~\citep{maetok} align the VAE latent to these features. A second family makes the tokenizer itself semantic. RAE~\citep{rae} pairs a frozen DINOv2 encoder with a trained decoder to obtain a representation latent for diffusion, VQRAE~\citep{vqrae} discretizes such a representation autoencoder, and TA-Tok~\citep{tar} and X-Omni~\citep{xomni} add vector quantization on top of a frozen SigLIP encoder. These semantic tokens speed up downstream generation and understanding, but because they are optimized for semantics rather than pixels, faithful reconstruction is hard, so TA-Tok and X-Omni rely on a separate generative de-tokenizer to render images. Recent analyses clarify which property actually matters. iREPA~\citep{irepa} finds that the \emph{spatial structure} of the target representation, not its global semantic accuracy, drives generation, and PAE~\citep{pae} finds that spatial-structure coherence and local continuity matter more than reconstruction fidelity. GEAR is consistent with these findings but reaches them differently. Instead of aligning to a fixed target offline, or trading away reconstruction by quantizing a semantic encoder, GEAR keeps a standard reconstruction tokenizer and lets the live AR model guide it end-to-end. This sharpens the \emph{patch-level} structure of the AR's features (\cref{fig:repr_analysis}) and re-organizes the tokenizer's index distribution toward predictability (\cref{fig:soft}), without sacrificing reconstruction.

\noindent\textbf{Toward end-to-end generative training.}
A growing line of work dissolves the two-stage boundary: REPA-E~\citep{repae} jointly trains the VAE and the diffusion model, and pixel-space transformers drop the tokenizer to generate on raw pixels~\citep{jit,pixeldit,pixelrepa,sensenovau1,tuna2}. Seen this way, the dividing line is not latent versus pixel but whether the pipeline is trained end-to-end, echoing detection's shift from the multi-stage R-CNN~\citep{rcnn} to single-stage, end-to-end detectors~\citep{yolo,detr}. The discrete VQ-AR setting is the hardest case, because the index is a non-differentiable $\arg\max$. Bridging it by sending the autoregressive gradient into the tokenizer through a straight-through estimator is unstable in our experiments (\cref{fig:teaser}) and, as the concurrent EOSTok~\citep{eostok} also reports, collapses the codebook. The cause is a conflict the next-token-prediction (NTP) loss creates once it can update the tokenizer: NTP rewards a token sequence that is easy to predict, which the tokenizer can trivially obtain by driving its index distribution to low entropy and emitting a few dominant codes, whereas faithful reconstruction demands a high-entropy, fully-used codebook. When the prediction loss reaches the tokenizer it wins this trade-off, collapsing codebook utilization and reconstruction fidelity. The same shortcut appears in continuous form as the latent-variance collapse REPA-E reports for VAEs: in both cases a downstream generation objective, given access to the tokenizer, sacrifices fidelity for predictability. EOSTok mitigates this by down-weighting the next-token-prediction loss to a coefficient of $0.1$ and adding an auxiliary pixel-space loss on its decoded AR predictions, which tacitly confirms that an undamped prediction loss on the tokenizer is harmful. GEAR instead removes the conflict at its source. The prediction loss never reaches the tokenizer: next-token prediction updates only the AR, while the tokenizer is shaped only by reconstruction and a representation-alignment signal carried through a differentiable soft assignment. Its codebook therefore stays broadly used rather than collapsing to a few dominant codes, and the AR trains and samples on exactly the discrete tokens it will see at inference.

\section{Method}
\label{sec:method}

\begin{figure}[t]
    \centering
    \includegraphics[width=\linewidth]{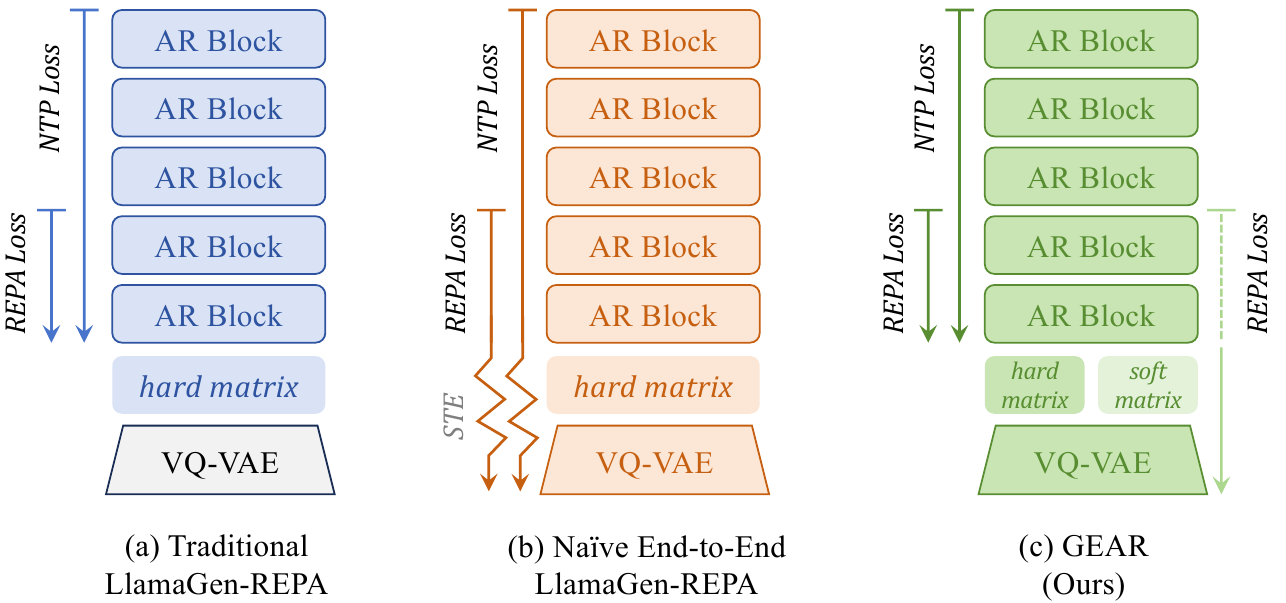}
    \caption{\textbf{Overview of GEAR.} \textbf{(a)} The conventional pipeline freezes a pretrained VQ-VAE and trains the AR model alone with the next-token-prediction (NTP) and REPA losses. \textbf{(b)} Naively making the pipeline end-to-end by passing AR gradients back into the tokenizer through the straight-through estimator (STE, drawn as the \textcolor{steorange}{zigzag arrows $\rightsquigarrow$}) is highly unstable and collapses (cf.\ \cref{tab:ablation}). \textbf{(c)} GEAR reads the per-position assignment both as a \emph{hard} (one-hot) and a \emph{soft} (temperature-scaled) matrix: the hard branch carries NTP and the hard REPA loss to update only the AR model, while the differentiable soft branch carries a REPA loss that bypasses the upper AR blocks and flows back (\textcolor{softgreen}{the dashed arrow $\dashrightarrow$}) to update only the tokenizer, giving a stable end-to-end guidance signal.}
    \label{fig:gear_overview}
\end{figure}

We present \textbf{GEAR}, a framework that trains the visual tokenizer and the autoregressive (AR) generator \emph{jointly and end-to-end}, so that the AR model can \emph{guide} the tokenizer toward a discrete index distribution that is easier to model causally. We first review the two components that GEAR unifies (\cref{sec:prelim}), and then describe how a dual hard/soft assignment couples them through representation alignment while keeping their optimization cleanly decoupled (\cref{sec:gear}). \cref{fig:gear_overview} contrasts GEAR with the conventional and naive end-to-end pipelines.

\subsection{Preliminaries}
\label{sec:prelim}

\myparagraph{Vector-quantized tokenization.}
A vector-quantized (VQ) tokenizer maps an image $\mathbf{x}\in\mathbb{R}^{H\times W\times 3}$ to a grid of discrete tokens. An encoder $\mathcal{E}$ produces a spatially down-sampled latent $\mathbf{Z}=\mathcal{E}(\mathbf{x})\in\mathbb{R}^{N\times d}$ with $N=h\times w$ positions (e.g., a $16\times$ down-sampling factor yields $h=w=16$ and $N=256$). Each latent vector $\mathbf{z}_i$ is quantized to its nearest entry in a learnable codebook $\mathcal{C}=\{\mathbf{c}_k\}_{k=1}^{K}\subset\mathbb{R}^{d}$,
\begin{equation}
q_i=\arg\min_{k}\;\lVert \mathbf{z}_i-\mathbf{c}_k\rVert_2^2,
\qquad
\hat{\mathbf{z}}_i=\mathbf{c}_{q_i},
\end{equation}
and a decoder $\mathcal{D}$ reconstructs $\hat{\mathbf{x}}=\mathcal{D}(\hat{\mathbf{Z}})$. The tokenizer is trained with a reconstruction term, a perceptual (LPIPS~\cite{lpips}) term, an adversarial term~\cite{vqgan}, a codebook entropy term that encourages full codebook usage, and a commitment term:
\begin{equation}
\label{eq:vqloss}
\mathcal{L}_{\mathrm{VQ}}
=\mathcal{L}_{\mathrm{rec}}
+0.1\,\mathcal{L}_{\mathrm{LPIPS}}
+0.1\,\mathcal{L}_{\mathrm{GAN}}
+0.05\,\mathcal{L}_{\mathrm{ent}}
+0.25\,\mathcal{L}_{\mathrm{commit}}.
\end{equation}
Because the $\arg\min$ assignment is non-differentiable, the gradient from $\mathcal{L}_{\mathrm{rec}}$ to the encoder is classically approximated by the straight-through estimator (STE)~\cite{vqvae}.

\myparagraph{Autoregressive image generation.}
Given the token grid $\mathbf{q}=(q_1,\dots,q_N)$ produced by the tokenizer and flattened in raster order, together with a generic condition $c$ (e.g., a class label or a text prompt), the AR generator models the joint distribution over these \emph{same} discrete indices causally,
\begin{equation}
p_\theta(\mathbf{q}\mid c)=\prod_{i=1}^{N}p_\theta\!\left(q_i \mid \mathbf{q}_{<i},\,c\right).
\end{equation}
The AR thus operates entirely at the level of discrete indices: its vocabulary is the set of $K$ codebook entries, and it both conditions on and predicts the index $q_i$ rather than the continuous code $\mathbf{c}_{q_i}$. For its input representation it does not reuse the VQ codebook $\mathcal{C}$. Instead, each index selects a row of the AR model's own learnable embedding table $\mathbf{E}\in\mathbb{R}^{K\times d}$, giving $\mathbf{u}_i=\mathbf{E}_{q_i}$. The condition is mapped to an embedding $\mathbf{e}_c$ and prepended, and the resulting sequence $\mathbf{S}=[\mathbf{e}_c,\mathbf{u}_1,\dots,\mathbf{u}_N]$ is processed by a stack of $L$ causal transformer blocks, producing hidden states $\mathbf{H}^{(\ell)}=(\mathbf{h}^{(\ell)}_1,\dots,\mathbf{h}^{(\ell)}_N)$ at every layer $\ell$. A linear head predicts, over the same $K$ indices, the distribution of the next token, trained by the next-token-prediction (NTP) loss
\begin{equation}
\label{eq:ntp}
\mathcal{L}_{\mathrm{NTP}}=-\sum_{i=1}^{N}\log p_\theta\!\left(q_i\mid \mathbf{q}_{<i},c\right).
\end{equation}
At inference the model samples tokens autoregressively starting from $c$, and the decoder $\mathcal{D}$ renders the completed grid into an image.

\myparagraph{Representation alignment.}
REPA~\cite{repa} accelerates generative training by aligning an intermediate hidden state of the generator with features from a frozen, pretrained vision encoder $f$ (e.g., DINOv2~\cite{dinov2}). Following its AR instantiation~\cite{llamagen-repa}, a lightweight projection head $g_\phi$ maps the hidden state at a chosen depth $\ell$ into the target space, and the two are aligned token-wise by maximizing cosine similarity:
\begin{equation}
\label{eq:align}
\mathcal{L}_{\mathrm{align}}\!\left(\mathbf{H}^{(\ell)}\right)
=-\frac{1}{N}\sum_{i=1}^{N}
\frac{\big\langle g_\phi(\mathbf{h}^{(\ell)}_i),\, f(\mathbf{x})_i\big\rangle}
{\lVert g_\phi(\mathbf{h}^{(\ell)}_i)\rVert_2\,\lVert f(\mathbf{x})_i\rVert_2}.
\end{equation}

\subsection{Guided End-to-End AutoRegression}
\label{sec:gear}

Conventional pipelines train the tokenizer first and \emph{freeze} it before training the AR generator. The discrete index distribution is therefore fixed by a pure reconstruction objective and is oblivious to whether the induced token sequence is easy to predict causally. GEAR removes this barrier: it optimizes the tokenizer and the AR model in a single end-to-end loop in which the AR model's representation objective \emph{guides} the tokenizer. The central difficulty is that the index assignment is discrete, so gradients cannot flow from the AR model back to the encoder. We find that naively bridging this gap with the STE is highly unstable (\cref{tab:ablation}). We instead introduce a differentiable guidance channel built from a soft assignment.

\myparagraph{Soft and hard assignments.}
For every position $i$ the tokenizer produces an assignment vector $\mathbf{A}_i$ over the $K$ codewords. Taking the simplest VQ tokenizer as the running example, we form it from the negative quantization distance to the codebook $\mathcal{C}$,
\begin{equation}
\mathbf{A}_{ik}=-\lVert \mathbf{z}_i-\mathbf{c}_k\rVert_2^2 ,\qquad \mathbf{A}\in\mathbb{R}^{N\times K}.
\end{equation}
For tokenizers with a different quantization rule, such as LFQ~\cite{magvitv2} or IBQ~\cite{ibq}, only the definition of $\mathbf{A}$ changes, and everything below is unchanged. We read $\mathbf{A}$ out in two complementary ways and map it onto the AR \emph{embedding} table $\mathbf{E}$ (not the codebook $\mathcal{C}$): the \emph{hard} read-out is the standard one-hot token lookup, whereas the \emph{soft} read-out is a temperature-controlled mixture of embeddings,
\begin{align}
\text{(hard)}\quad & \mathbf{u}^{\mathrm{h}}_i=\sum_{k}\mathds{1}\!\left[k=\arg\max_{k'}\mathbf{A}_{ik'}\right]\mathbf{e}_k=\mathbf{E}_{q_i},\\
\text{(soft)}\quad & \mathbf{u}^{\mathrm{s}}_i=\sum_{k}\boldsymbol{\pi}_{ik}\,\mathbf{e}_k=\boldsymbol{\pi}_i\mathbf{E},\qquad
\boldsymbol{\pi}_i=\mathrm{softmax}\!\left(\mathbf{A}_i/\tau\right),
\end{align}
where $\tau>0$ is a guidance temperature and $\mathbf{e}_k$ is the $k$-th row of $\mathbf{E}$. Hence the discrete index is decided by the codebook $\mathcal{C}$, but the vector fed to the AR model is read from its own embedding table $\mathbf{E}$. The soft read-out $\mathbf{u}^{\mathrm{s}}_i$ is a fully differentiable function of the assignment $\boldsymbol{\pi}_i$ (hence of the encoder $\mathcal{E}$ and codebook $\mathcal{C}$) and of $\mathbf{E}$, and $\mathbf{u}^{\mathrm{s}}_i\!\to\!\mathbf{u}^{\mathrm{h}}_i$ as $\tau\!\to\!0$. We use the hard read-out to define the discrete sequence the AR model must ultimately predict, and the soft read-out as a differentiable surrogate that carries gradients back into the original tokenizer.

\myparagraph{Dual-branch forward.}
We build two condition-prefixed sequences,
\begin{equation}
\mathbf{S}^{\mathrm{h}}=\big[\mathbf{e}_c,\,\mathbf{u}^{\mathrm{h}}_1,\dots,\mathbf{u}^{\mathrm{h}}_N\big],
\qquad
\mathbf{S}^{\mathrm{s}}=\big[\mathbf{e}_c,\,\mathbf{u}^{\mathrm{s}}_1,\dots,\mathbf{u}^{\mathrm{s}}_N\big],
\end{equation}
and feed both through the causal AR backbone. The hard branch passes through the non-differentiable $\arg\max$, so it carries no gradient to the tokenizer ($\mathcal{E}$, $\mathcal{C}$) and matches the discrete tokens used at inference. The soft branch instead keeps a live, differentiable connection to the tokenizer through $\boldsymbol{\pi}$. The hard branch is run to the final layer to produce next-token logits, whereas the soft branch only needs to reach the alignment depth $\ell$ and is \emph{truncated} there, adding negligible compute.

\myparagraph{Representation guidance.}
At depth $\ell$ both branches emit hidden states, $\mathbf{H}^{(\ell),\mathrm{h}}$ and $\mathbf{H}^{(\ell),\mathrm{s}}$, and we apply the alignment loss of \cref{eq:align} to each against the same target features $f(\mathbf{x})$:
\begin{equation}
\mathcal{L}^{\mathrm{h}}_{\mathrm{align}}=\mathcal{L}_{\mathrm{align}}\!\left(\mathbf{H}^{(\ell),\mathrm{h}}\right),
\qquad
\mathcal{L}^{\mathrm{s}}_{\mathrm{align}}=\mathcal{L}_{\mathrm{align}}\!\left(\mathbf{H}^{(\ell),\mathrm{s}}\right).
\end{equation}
The two alignment terms play distinct roles. $\mathcal{L}^{\mathrm{h}}_{\mathrm{align}}$ regularizes the \emph{generator} exactly as in REPA, operating on the hard, inference-time tokens. $\mathcal{L}^{\mathrm{s}}_{\mathrm{align}}$ is the \emph{guidance signal}: because $\mathbf{u}^{\mathrm{s}}$ is differentiable, its gradient flows through the (otherwise fixed) AR backbone back to the encoder and the codebook, guiding the tokenizer to reshape its assignment so that the induced tokens are easier for the AR model to predict and to align with DINOv2.

\myparagraph{Decoupled optimization.}
GEAR keeps the two modules' updates disjoint. The tokenizer parameters $\theta_{\mathrm{tok}}=\{\mathcal{E},\mathcal{C},\mathcal{D}\}$ are updated by the VQ objective (\cref{eq:vqloss}) together with the soft guidance term, while the AR parameters $\theta_{\mathrm{AR}}=\{\text{transformer},\,\text{embedding }\mathbf{E},\,\text{condition embedding},\,\text{head},\,g_\phi\}$ are updated by NTP together with the hard alignment term:
\begin{align}
\label{eq:update_tok}
\theta_{\mathrm{tok}} &\;\leftarrow\; \theta_{\mathrm{tok}}-\eta\,\nabla_{\theta_{\mathrm{tok}}}\big(\mathcal{L}_{\mathrm{VQ}}+\lambda\,\mathcal{L}^{\mathrm{s}}_{\mathrm{align}}\big),\\
\label{eq:update_ar}
\theta_{\mathrm{AR}} &\;\leftarrow\; \theta_{\mathrm{AR}}-\eta\,\nabla_{\theta_{\mathrm{AR}}}\big(\mathcal{L}_{\mathrm{NTP}}+\lambda\,\mathcal{L}^{\mathrm{h}}_{\mathrm{align}}\big),
\end{align}
with a single alignment coefficient $\lambda$. Concretely, the soft guidance gradient updates only the tokenizer (the AR backbone, embedding $\mathbf{E}$ and projector $g_\phi$ are held fixed for this term), while the non-differentiable $\arg\max$ on the hard branch ensures that NTP and $\mathcal{L}^{\mathrm{h}}_{\mathrm{align}}$ update only the AR model. The encoder's \emph{end-to-end} guidance therefore arrives through the differentiable soft assignment rather than the unstable STE, on top of the standard tokenizer objective $\mathcal{L}_{\mathrm{VQ}}$. \cref{eq:update_tok,eq:update_ar} make the \emph{guided} nature of GEAR explicit: a shared alignment target trains the generator on the hard, inference-time tokens while simultaneously steering the tokenizer through the soft, differentiable ones. We summarize the full details of procedure as shown in \cref{alg:gear}.

\begin{algorithm}[t]
\caption{One GEAR training step}
\label{alg:gear}
\begin{algorithmic}[1]
\Require image $\mathbf{x}$, condition $c$, temperature $\tau$, coefficient $\lambda$, depth $\ell$
\State $\mathbf{Z}\gets\mathcal{E}(\mathbf{x})$; \quad $\mathbf{A}_{ik}\gets-\lVert\mathbf{z}_i-\mathbf{c}_k\rVert_2^2$ \Comment{assignment to codebook $\mathcal{C}$}
\State $\hat{\mathbf{z}}\gets\mathbf{c}_{\arg\max\mathbf{A}}$;\quad $\hat{\mathbf{x}}\gets\mathcal{D}(\hat{\mathbf{z}})$;\quad evaluate $\mathcal{L}_{\mathrm{VQ}}$ \Comment{VQ reconstruction}
\State $\mathbf{u}^{\mathrm{h}}\gets\mathbf{E}_{\arg\max\mathbf{A}}$;\quad $\mathbf{u}^{\mathrm{s}}\gets\mathrm{softmax}(\mathbf{A}/\tau)\,\mathbf{E}$ \Comment{AR hard / soft embeddings}
\State forward $\mathbf{S}^{\mathrm{h}}=[\mathbf{e}_c,\mathbf{u}^{\mathrm{h}}]$ to layer $L$, and $\mathbf{S}^{\mathrm{s}}=[\mathbf{e}_c,\mathbf{u}^{\mathrm{s}}]$ to layer $\ell$
\State evaluate $\mathcal{L}_{\mathrm{NTP}},\,\mathcal{L}^{\mathrm{h}}_{\mathrm{align}}$ (hard branch) and $\mathcal{L}^{\mathrm{s}}_{\mathrm{align}}$ (soft branch)
\State $\theta_{\mathrm{tok}}\!\gets\!\theta_{\mathrm{tok}}-\eta\nabla_{\theta_{\mathrm{tok}}}(\mathcal{L}_{\mathrm{VQ}}+\lambda\mathcal{L}^{\mathrm{s}}_{\mathrm{align}})$ \Comment{guidance $\to$ tokenizer}
\State $\theta_{\mathrm{AR}}\!\gets\!\theta_{\mathrm{AR}}-\eta\nabla_{\theta_{\mathrm{AR}}}(\mathcal{L}_{\mathrm{NTP}}+\lambda\mathcal{L}^{\mathrm{h}}_{\mathrm{align}})$ \Comment{NTP $\to$ AR}
\end{algorithmic}
\end{algorithm}

\section{Experiment}
\label{sec:experiment}

\subsection{Experimental Setup}
\label{sec:setup}

\myparagraph{Datasets and metrics.}
We study class-conditional image generation on ImageNet-1K at $256\times256$ resolution. For generation quality we report the generation FID (gFID), spatial FID (sFID), Inception Score (IS), Precision (Prec.) and Recall (Rec.), computed on 50K samples following the standard ADM evaluation protocol. To probe the co-trained tokenizer itself, we additionally report reconstruction FID (rFID), PSNR and SSIM. Unless stated otherwise, generation metrics are reported without classifier-free guidance (CFG). For text-to-image generation we first report a strictly controlled comparison on GPIC~\citep{gpic}, where all methods are trained for a single epoch on the same $100$M-image set. Following the official toolkit, on GPIC we report the Fr\'echet distance in DINOv2 feature space (FD-DINOv2, abbreviated FDD), Precision (Prec.), Recall (Rec.), Density (Dens.), Coverage (Cov.) and MMD. We additionally report the GenEval~\citep{geneval} and DPG-Bench~\citep{dpgbench} benchmarks, together with the CLIP Score and FID computed on the COCO 2017 validation set ($5$k image-text pairs). The CLIP Score is computed with the \texttt{openai/clip-vit-base-patch32} model.

\myparagraph{Training protocol.}
We compare against LlamaGen~\citep{llamagen}, which couples a VQ tokenizer with a Llama-style causal transformer, and LlamaGen-REPA~\citep{llamagen-repa}, which adds representation alignment to the AR model. Our LlamaGen-REPA and GEAR runs both start from the same warm-up tokenizer, a brief fine-tune that recovers the GAN discriminator omitted by public VQ tokenizers, following REPA-E~\citep{repae}, and barely changes reconstruction. From this shared start the two regimes differ. In our ablations and representation analysis, GEAR fine-tunes the tokenizer end-to-end jointly with the AR (\cref{eq:update_tok,eq:update_ar}), whereas LlamaGen-REPA keeps it frozen. For the main results (\cref{tab:comparison_detail,tab:gpic_t2i_performance,tab:t2i_benchmarks}) the long schedules make full joint training impractical, so GEAR instead freezes its end-to-end-tuned tokenizer, produced by the $400$k-step joint run, and trains a fresh AR on top, exactly as LlamaGen-REPA trains on the frozen warm-up tokenizer. The two methods thus share an identical AR and training budget and differ only in the frozen tokenizer, so any gain isolates the tokenizer's contribution and shows that it transfers: the end-to-end-improved tokenizer can be dropped into a standard frozen-tokenizer pipeline without paying the end-to-end cost. To avoid confounding with model size, each scale uses its own matched tokenizer. Optimization and per-experiment hyperparameters are deferred to \cref{app:ablation_configs,app:t2i}.

\myparagraph{Text-to-image instantiation.}
For text-to-image generation the condition is a sequence of text tokens from a Qwen3-1.7B~\citep{qwen3} text encoder, following GPIC~\citep{gpic}. The model is a strict autoregressor over the concatenation of text and image tokens, trained from scratch on the $100$M-image GPIC corpus. Its backbone, hybrid stream design and text-branch initialization are detailed in \cref{app:t2i}. GEAR and LlamaGen-REPA share this encoder and training and differ only in the frozen tokenizer. We report the GPIC toolkit metrics (\cref{tab:gpic_t2i_performance}) and standard text-to-image benchmarks (\cref{tab:t2i_benchmarks}).

\subsection{Main Results}
\label{sec:main_results}

\myparagraph{Class-conditional ImageNet.}
\cref{tab:comparison_detail} compares GEAR with representative latent diffusion models (DiT~\citep{dit}, SiT~\citep{sit}, MDT~\citep{mdt,mdtv2}) and autoregressive generators (LlamaGen, LlamaGen-REPA) on ImageNet $256\times256$, where the $111$M/$343$M/$775$M rows of the AR models denote the B/L/XL variants. At a matched $300$ epochs and parameter count, GEAR clearly improves over LlamaGen-REPA: with CFG, gFID drops from $6.00$ to $4.95$ ($111$M), $3.15$ to $2.95$ ($343$M) and $2.68$ to $2.52$ ($775$M), with consistently higher IS. End-to-end training also improves the tokenizer itself, with the per-scale behavior analyzed in the model-size ablation (\cref{tab:different_size}).

\myparagraph{Text-to-image generation.}
We further apply GEAR to text-to-image synthesis, where the condition $c$ in \cref{eq:ntp} is a text prompt. Our main, strictly controlled comparison is on GPIC~\cite{gpic} (\cref{tab:gpic_t2i_performance}): every model uses the same Qwen3-1.7B~\cite{qwen3} text encoder and is trained for a single epoch on the same $100$M-image corpus, so differences reflect the method alone. Two observations stand out. First, at matched budgets GEAR consistently outperforms LlamaGen-REPA: across $50$k/$100$k/$200$k/$390$k steps it lowers the GPIC FDD with CFG to $256.9/177.4/138.0/115.3$ (vs.\ $279.6/198.6/153.5/127.9$ for LlamaGen-REPA), mirroring the class-conditional gains. Second, autoregressive models converge far faster than the diffusion baseline: LlamaGen-REPA already surpasses the $390$k-step JiT-GPIC result (FDD $204.0$) after only $100$k steps (FDD $198.6$). We further report the same GEAR and LlamaGen-REPA models on the GenEval~\cite{geneval} and DPG-Bench~\cite{dpgbench} benchmarks, together with the CLIP Score and FID on the COCO 2017 validation set (\cref{tab:t2i_benchmarks}). In contrast to the distribution-level metrics (FDD, FID), GenEval, DPG-Bench and the CLIP Score probe \emph{prompt adherence}, i.e.\ instruction following. The absolute scores are modest here for two reasons: the single-epoch model has not yet converged, and we train at $256$ resolution but resize the samples to $512$ for evaluation. As elsewhere, our aim is not to top these leaderboards but to compare methods under a controlled setting and isolate the effect of GEAR's end-to-end training. A full classifier-free guidance sweep on these benchmarks, with a discussion of this convergence behavior, is deferred to \cref{app:t2i_cfg}.

\begin{figure}[t]
    \centering
    \includegraphics[width=\linewidth]{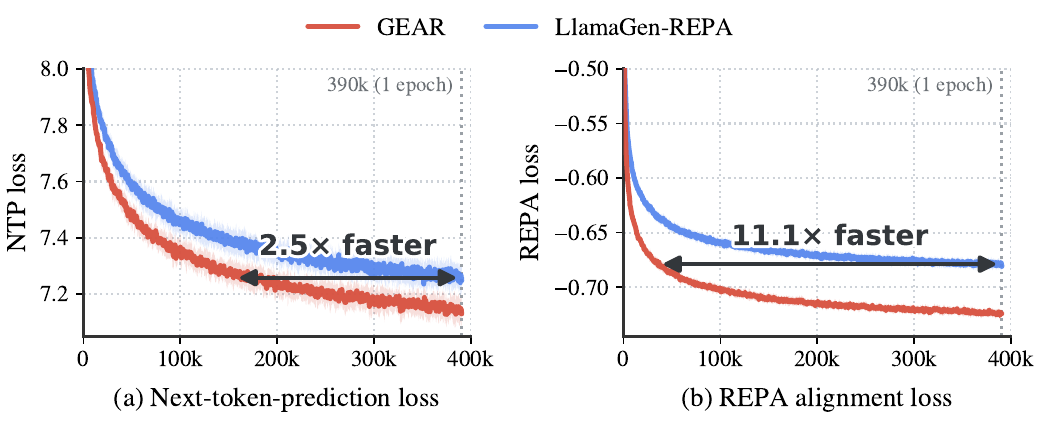}
    \caption{\textbf{Training dynamics on GPIC.} NTP loss (a) and REPA alignment loss (b) over the single epoch ($390$k steps). We train an AR on \textcolor{gearred}{GEAR}'s end-to-end-improved tokenizer and on the \textcolor{reprablue}{original} tokenizer. Both are kept frozen here, so the two runs differ only in tokenizer quality. To reach the baseline's final loss, GEAR is $2.5\times$ faster on NTP and $11.1\times$ faster on REPA alignment.}
    \label{fig:trainloss}
\end{figure}

\cref{fig:trainloss} shows where these GPIC gains come from. The tokenizer is frozen in both runs, so the two differ only in its quality. Even so, the AR trained on GEAR's tokenizer converges faster on both objectives: on GPIC it reaches the baseline's final REPA-alignment loss $11.1\times$ faster and its NTP loss $2.5\times$ faster. This acceleration is a property that the tokenizer carries over from end-to-end training: its tokens form a more predictable grid, so a freshly trained AR fits them with much less compute and reaches stronger patch-level DINOv2 alignment (\cref{fig:repr_analysis}).

\begin{table}[t]
    \caption{\textbf{System-Level Comparison on Class-Conditional ImageNet $256\times256$.} For the AR models, $111$M/$343$M/$775$M denote the B/L/XL variants, and $^{*}$ marks inference at $384$ resolution evaluated at $256$. \textcolor{gray}{ImageNet val.} is the real-data reference (an oracle floor). It is CFG-agnostic and shown once under the w/o-CFG columns. The w/ CFG results for LlamaGen-REPA and GEAR use a guidance scale of $1.5$. The rows are grouped by (epochs, params), and within each group the best value across the three models is in \textbf{bold}.}
    \centering
    \setlength{\tabcolsep}{0.8mm}{
    \begin{tabularx}{\linewidth}{l|c|c|*{5}{Y}|*{5}{Y}}
    \toprule
     \multirow{2}{*}{\textbf{Method}} & \multirow{2}{*}{\textbf{Epochs}} & \multirow{2}{*}{\textbf{Params}} & \multicolumn{5}{c|}{\textbf{Generation w/o CFG}} & \multicolumn{5}{c}{\textbf{Generation w/ CFG}} \\
    \cmidrule(lr){4-8} \cmidrule(lr){9-13}
     &  &  & \textbf{gFID}$\downarrow$ & \textbf{sFID}$\downarrow$ & \textbf{IS}$\uparrow$ & \textbf{Prec.}$\uparrow$ & \textbf{Rec.}$\uparrow$ & \textbf{gFID}$\downarrow$ & \textbf{sFID}$\downarrow$ & \textbf{IS}$\uparrow$ & \textbf{Prec.}$\uparrow$ & \textbf{Rec.}$\uparrow$ \\
    \midrule
    \rowcolor{gray!8} \textcolor{gray}{ImageNet val.~(ref.)} & \textcolor{gray}{-} & \textcolor{gray}{-} & \textcolor{gray}{1.78} & \textcolor{gray}{-} & \textcolor{gray}{236.9} & \textcolor{gray}{0.75} & \textcolor{gray}{0.67} & \textcolor{gray}{-} & \textcolor{gray}{-} & \textcolor{gray}{-} & \textcolor{gray}{-} & \textcolor{gray}{-} \\
    \midrule
    
    \rowcolor{blue!8}\multicolumn{13}{c}{\textbf{Latent Diffusion Models (LDM)}} \\
    \midrule
    DiT~\cite{dit} & 1400 & 675M & 9.62 & 6.85  & 121.5  & 0.67 & 0.67 & 2.27 & 4.60 & 278.2 & \textbf{0.83} & 0.57 \\
    SiT~\cite{sit} & 1400 & 675M & 8.61 & 6.32  & 131.7  & 0.68 & 0.67 & 2.06 & 4.50 & 270.3 & 0.82 & 0.59 \\
    MDT~\cite{mdt} & 1300 & 675M & 6.23 & 5.23  & 143.0  & 0.71 & 0.65 & 1.79 & 4.57 & 283.0 & 0.81 & 0.61 \\
    MDTv2~\cite{mdtv2} & 1080 & 675M & - & - & - & - & - & 1.58 & 4.52 & \textbf{314.7} & 0.79 & 0.65 \\
    SiT-REPA~\cite{repa} & 800 & 675M & 5.90 & 5.73 & 157.8 & 0.70 & \textbf{0.69} & 1.90 & 4.48 & 297.5 & 0.82 & 0.60 \\
    REPA-E~\cite{repae} & 800 & 675M & \textbf{1.69} & \textbf{4.17} & \textbf{219.3} & \textbf{0.77} & 0.67 & \textbf{1.12} & \textbf{4.09} & 302.9 & 0.79 & \textbf{0.66} \\
    \midrule
    
    \rowcolor{blue!8}\multicolumn{13}{c}{\text{\textbf{AutoRegressive (AR)}}} \\
    \midrule

    
    LlamaGen~\cite{llamagen} & 300 & 111M & 26.26 & 9.21 & 48.07 & 0.59 & 0.61 & 8.73 & 7.66 & 129.60 & 0.75 & \textbf{0.53} \\
    LlamaGen-REPA~\cite{llamagen-repa} & 300 & 111M & 20.16 & 7.03 & 60.66 & 0.64 & \textbf{0.62} & 6.00 & 5.77 & 144.98 & 0.79 & \textbf{0.53} \\
    \rowcolor{red!8} \textbf{GEAR (Ours)} & 300 & 111M & \textbf{16.96} & \textbf{5.55} & \textbf{67.62} & \textbf{0.67} & \textbf{0.62} & \textbf{4.95} & \textbf{5.08} & \textbf{166.13} & \textbf{0.82} & 0.52 \\
    \midrule
    LlamaGen~\cite{llamagen} & 300 & 343M & 13.45 & 8.32 & 82.28 & 0.65 & 0.63 & 4.07 & 8.15 & 198.50 & 0.80 & 0.55 \\
    LlamaGen-REPA~\cite{llamagen-repa} & 300 & 343M & 12.70 & 6.09 & 89.16 & 0.67 & \textbf{0.65} & 3.15 & \textbf{5.26} & 208.14 & 0.80 & \textbf{0.57} \\
    \rowcolor{red!8} \textbf{GEAR (Ours)} & 300 & 343M & \textbf{8.66} & \textbf{5.04} & \textbf{107.52} & \textbf{0.71} & 0.63 & \textbf{2.95} & 5.40 & \textbf{239.75} & \textbf{0.84} & 0.54 \\
    \midrule
    LlamaGen~\cite{llamagen} & 300 & 775M & 15.54$^*$ & 7.04$^*$ & 79.15$^*$ & 0.61$^*$ & \textbf{0.68}$^*$ & 3.47$^*$ & 5.81$^*$ & 194.44$^*$ & 0.76$^*$ & \textbf{0.60}$^*$ \\
    LlamaGen-REPA~\cite{llamagen-repa} & 300 & 775M & 8.20 & 5.56 & 115.06 & 0.70 & 0.65 & 2.68 & 5.45 & 232.15 & 0.81 & 0.58 \\
    \rowcolor{red!8} \textbf{GEAR (Ours)} & 300 & 775M & \textbf{6.76} & \textbf{4.96} & \textbf{129.00} & \textbf{0.72} & 0.65 & \textbf{2.52} & \textbf{5.14} & \textbf{262.94} & \textbf{0.84} & 0.55 \\
    \midrule
    \midrule
    LlamaGen-REPA~\cite{llamagen-repa} & 800 & 111M & 19.14 & 7.52 & 64.27 & 0.64 & \textbf{0.62} & 5.30 & 5.83 & 158.87 & 0.79 & \textbf{0.54} \\
    \rowcolor{red!8} \textbf{GEAR (Ours)} & 800 & 111M & \textbf{14.98} & \textbf{5.27} & \textbf{73.63} & \textbf{0.68} & \textbf{0.62} & \textbf{4.35} & \textbf{4.96} & \textbf{180.85} & \textbf{0.82} & 0.52 \\
    \midrule
    LlamaGen-REPA~\cite{llamagen-repa} & 800 & 343M & 10.44 & 5.85 & 100.54 & 0.68 & \textbf{0.65} & 2.92 & 5.40 & 215.99 & 0.80 & \textbf{0.57} \\
    \rowcolor{red!8} \textbf{GEAR (Ours)} & 800 & 343M & \textbf{8.61} & \textbf{5.07} & \textbf{109.80} & \textbf{0.71} & 0.64 & \textbf{2.72} & \textbf{5.22} & \textbf{240.30} & \textbf{0.83} & 0.55 \\
    \midrule
    LlamaGen-REPA~\cite{llamagen-repa} & 800 & 775M & 7.46 & 5.74 & 124.49 & 0.71 & \textbf{0.65} & 2.57 & 5.36 & 236.27 & 0.81 & \textbf{0.58} \\
    \rowcolor{red!8} \textbf{GEAR (Ours)} & 800 & 775M & \textbf{6.28} & \textbf{4.99} & \textbf{136.73} & \textbf{0.73} & \textbf{0.65} & \textbf{2.45} & \textbf{5.05} & \textbf{254.27} & \textbf{0.83} & \textbf{0.58} \\

    \bottomrule
    \end{tabularx}
    }
    \label{tab:comparison_detail}
\end{table}

\begin{table}[!t]
    \caption{\textbf{System-Level Comparison on Text-to-Image Generation (GPIC).} All models share the same Qwen3-1.7B text encoder and are evaluated on the GPIC $50$k test set with the official toolkit~\cite{gpic}. FDD is the Fr\'echet distance in DINOv2 feature space (FD-DINOv2). $390$k steps corresponds to roughly one epoch over the $100$M-image training set at a global batch size of $256$. The w/ CFG columns use a guidance scale of $1.75$. Rows are grouped by training steps, and within each group the better of LlamaGen-REPA and GEAR is in \textbf{bold}.}
    \centering
    \setlength{\tabcolsep}{0.8mm}{
    \begin{tabularx}{\linewidth}{l|c|*{5}{Y}|*{5}{Y}}
    \toprule
     \multirow{2}{*}{\textbf{Method}}  & \multirow{2}{*}{\textbf{Steps}} & \multicolumn{5}{c|}{\textbf{Generation w/o CFG}} & \multicolumn{5}{c}{\textbf{Generation w/ CFG}} \\
    \cmidrule(lr){3-7} \cmidrule(lr){8-12}
     & & \textbf{FDD}$\downarrow$ & \textbf{Prec.}$\uparrow$ & \textbf{Rec.}$\uparrow$ & \textbf{Cov.}$\uparrow$ & \textbf{MMD}$\downarrow$ & \textbf{FDD}$\downarrow$ & \textbf{Prec.}$\uparrow$ & \textbf{Rec.}$\uparrow$ & \textbf{Cov.}$\uparrow$ & \textbf{MMD}$\downarrow$ \\
    \midrule
     JiT-GPIC-1.1B~\cite{gpic} & 390k & - & - & - & - & - & 204.0 & 0.91 & 0.53 & 0.80 & - \\
    \midrule
     LlamaGen-REPA-1.0B~\cite{llamagen-repa} & 50k & 414.8 & \textbf{0.86} & 0.33 & 0.52 & 0.59 & 279.6 & \textbf{0.87} & 0.46 & 0.67 & 0.35 \\
     \rowcolor{red!8} \textbf{GEAR-1.0B (Ours)} & 50k & \textbf{381.6} & \textbf{0.86} & \textbf{0.35} & \textbf{0.56} & \textbf{0.55} & \textbf{256.9} & \textbf{0.87} & \textbf{0.50} & \textbf{0.69} & \textbf{0.32} \\
    \midrule
     LlamaGen-REPA-1.0B~\cite{llamagen-repa} & 100k & 319.2 & 0.86 & 0.44 & 0.63 & 0.44 & 198.6 & 0.89 & 0.61 & 0.77 & 0.23 \\
     \rowcolor{red!8} \textbf{GEAR-1.0B (Ours)} & 100k & \textbf{281.3} & \textbf{0.88} & \textbf{0.50} & \textbf{0.67} & \textbf{0.38} & \textbf{177.4} & \textbf{0.91} & \textbf{0.65} & \textbf{0.78} & \textbf{0.20} \\
    \midrule
     LlamaGen-REPA-1.0B~\cite{llamagen-repa} & 200k & 261.4 & 0.88 & 0.56 & 0.68 & 0.34 & 153.5 & 0.90 & 0.72 & 0.82 & 0.17 \\
     \rowcolor{red!8} \textbf{GEAR-1.0B (Ours)} & 200k & \textbf{230.0} & \textbf{0.89} & \textbf{0.61} & \textbf{0.73} & \textbf{0.30} & \textbf{138.0} & \textbf{0.91} & \textbf{0.75} & \textbf{0.84} & \textbf{0.15} \\
    \midrule
     LlamaGen-REPA-1.0B~\cite{llamagen-repa} & 390k & 228.9 & 0.89 & 0.63 & 0.74 & 0.30 & 127.9 & \textbf{0.92} & 0.78 & 0.85 & 0.14 \\
     \rowcolor{red!8} \textbf{GEAR-1.0B (Ours)} & 390k & \textbf{200.9} & \textbf{0.90} & \textbf{0.66} & \textbf{0.77} & \textbf{0.25} & \textbf{115.3} & \textbf{0.92} & \textbf{0.80} & \textbf{0.88} & \textbf{0.12} \\
    \bottomrule
    \end{tabularx}
    }
    \label{tab:gpic_t2i_performance}
\end{table}

\begin{table}[!t]
    \caption{\textbf{Text-to-Image Evaluation on Broader Benchmarks.} The top block lists representative text-to-image systems for context. They differ in data, architecture, training schedule, inference (e.g.,\ CFG scale), and tokenizer, and we report their published with-CFG GenEval (short prompts) and DPG-Bench. The bottom block is our controlled comparison: GEAR and LlamaGen-REPA ($1.0$B) share the same Qwen3-1.7B text encoder and GPIC training and differ only in the (end-to-end-tuned) frozen tokenizer. GenEval~\cite{geneval} is reported under \emph{short} (original) and \emph{long} (LLM-refined) prompts, and the CLIP Score and FID are on the COCO 2017 validation set ($5$k pairs). Within the controlled comparison the better of the two is in \textbf{bold}, and ``--'' marks numbers not reported.}
    \centering
    \setlength{\tabcolsep}{0.8mm}{
    \begin{tabularx}{\linewidth}{l|*{5}{Y}|*{5}{Y}}
    \toprule
     \multirow{3}{*}{\textbf{Method}} & \multicolumn{5}{c|}{\textbf{Generation w/o CFG}} & \multicolumn{5}{c}{\textbf{Generation w/ CFG}} \\
    \cmidrule(lr){2-6} \cmidrule(lr){7-11}
     & \multicolumn{2}{c}{\textbf{GenEval}$\uparrow$} & \multirow{2}{*}{\makecell{\textbf{DPG-}\\\textbf{Bench}$\uparrow$}} & \multirow{2}{*}{\makecell{\textbf{CLIP}\\\textbf{Score}$\uparrow$}} & \multirow{2}{*}{\textbf{FID}$\downarrow$} & \multicolumn{2}{c}{\textbf{GenEval}$\uparrow$} & \multirow{2}{*}{\makecell{\textbf{DPG-}\\\textbf{Bench}$\uparrow$}} & \multirow{2}{*}{\makecell{\textbf{CLIP}\\\textbf{Score}$\uparrow$}} & \multirow{2}{*}{\textbf{FID}$\downarrow$} \\
    \cmidrule(lr){2-3} \cmidrule(lr){7-8}
     & short & long & & & & short & long & & & \\
    \midrule
    \rowcolor{blue!8} \multicolumn{11}{c}{\textbf{Other systems} \,(different data, architecture, schedule, inference, and tokenizer)} \\
    \midrule
    PixArt-$\alpha$-0.6B~\cite{pixarta} & - & - & - & - & - & 0.48 & - & 71.11 & - & - \\
    Chameleon-7B~\cite{chameleon} & - & - & - & - & - & 0.39 & - & - & - & - \\
    \midrule
    \rowcolor{blue!8} \multicolumn{11}{c}{\textbf{Controlled comparison} \,(same data, architecture, schedule, and inference, differing only in the tokenizer)} \\
    \midrule
    LlamaGen-REPA-1.0B~\cite{llamagen-repa} & 0.074 & 0.218 & \textbf{55.386} & 27.12 & 30.04 & 0.272 & 0.419 & 70.363 & 30.80 & 27.99 \\
    \rowcolor{red!8} \textbf{GEAR-1.0B (Ours)} & \textbf{0.086} & \textbf{0.227} & 55.369 & \textbf{27.29} & \textbf{29.66} & \textbf{0.334} & \textbf{0.478} & \textbf{72.881} & \textbf{31.39} & \textbf{25.74} \\
    \bottomrule
    \end{tabularx}
    }
    \label{tab:t2i_benchmarks}
\end{table}

\subsection{Representation Analysis}
\label{sec:repr_analysis}

\begin{figure}[!t]
    \centering
    \includegraphics[width=\linewidth]{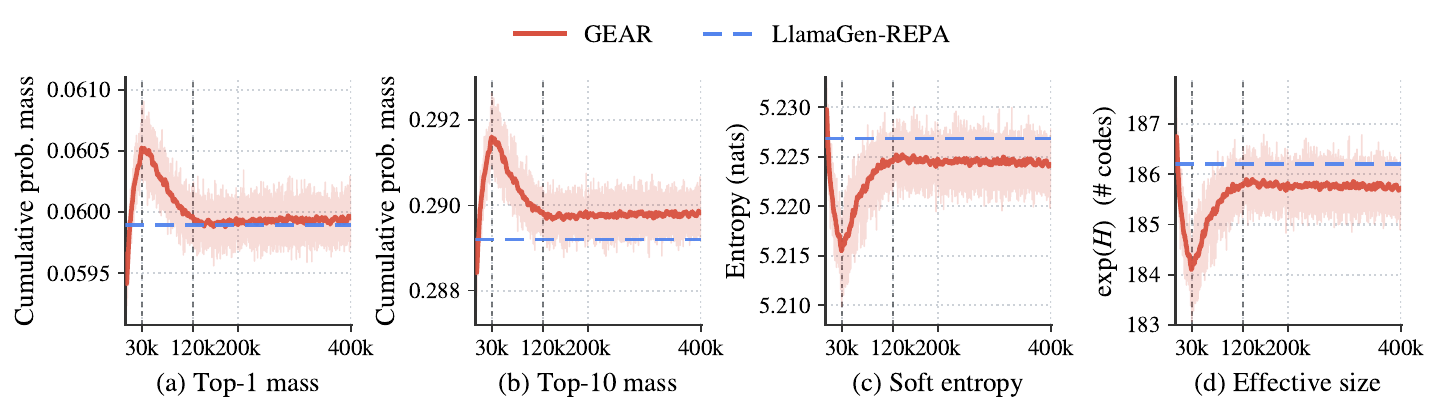}
    \caption{\textbf{Codebook usage during end-to-end fine-tuning.} Over GEAR's joint fine-tuning on ImageNet ($\tau{=}0.1$), we track the cumulative top-$1$/$10$ assignment mass (a--c), the usage entropy in nats (d), and the effective codebook size $\exp(H)$ (e). The blue dashed line marks the warm-up tokenizer, the start of fine-tuning that LlamaGen-REPA uses. The distribution sharpens rapidly, peaks in concentration near $30$k steps, then relaxes and converges (after ${\sim}120$k) to a state more concentrated and lower-entropy than the warm-up.}
    \label{fig:soft}
\end{figure}

\begin{figure*}[!t]
    \centering
    \includegraphics[width=\linewidth]{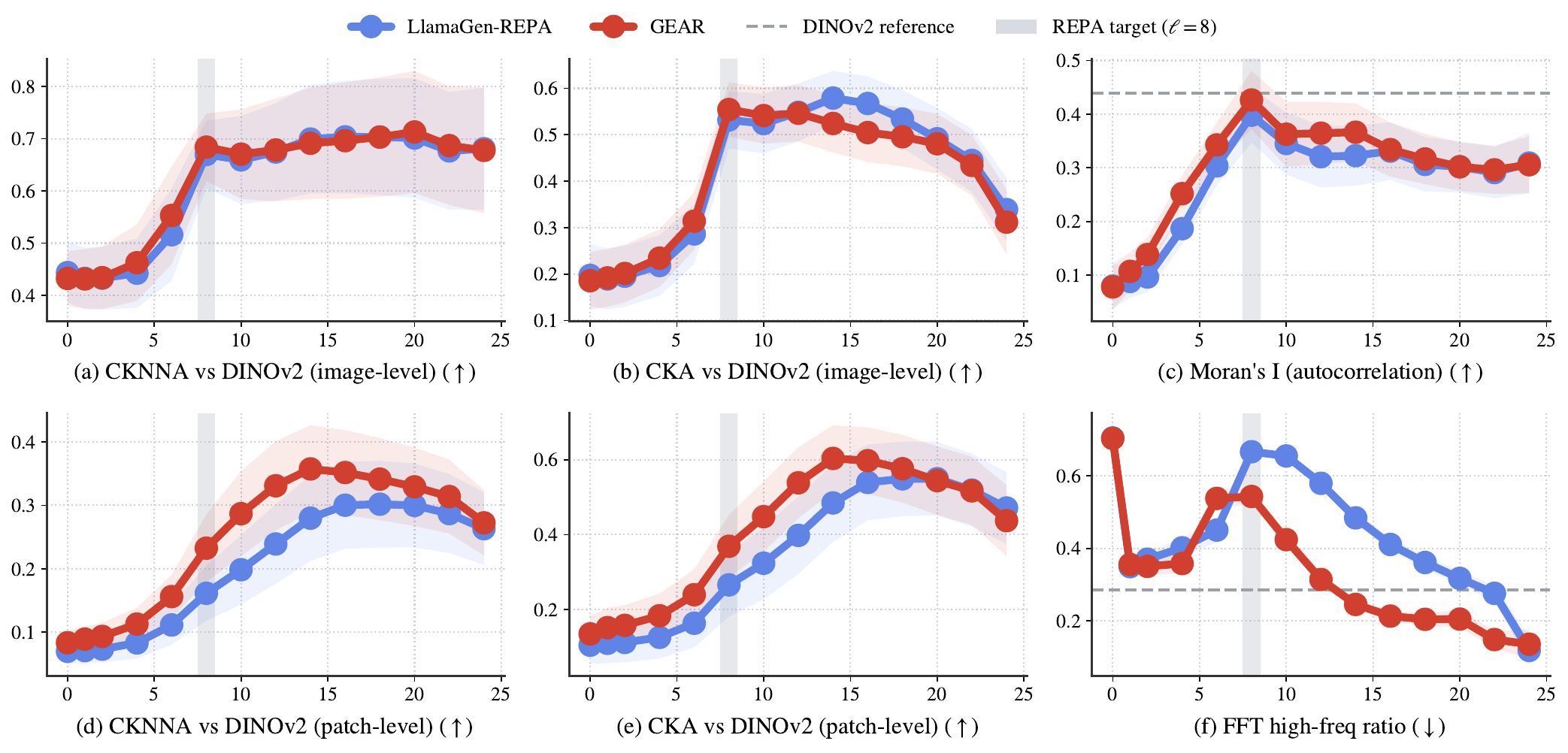}
    \caption{\textbf{AR-feature alignment to DINOv2 across depth.} \textcolor{gearred}{GEAR-L} vs.\ \textcolor{reprablue}{LlamaGen-REPA-L}, both aligning the $8$-th layer to DINOv2 (\colorbox{black!10}{gray band}) and trained for $400$k steps. The horizontal axis is the AR layer index, where layer $0$ is the output of the AR embedding layer. We take the \emph{raw} per-layer hidden states (before the REPA projection head) and measure alignment to DINOv2 via CKNNA~\cite{cknna} and CKA~\cite{cka} at the image level (a,b) and patch level (d,e), together with Moran's~$I$~\cite{moran} (c) and the FFT high-frequency ratio (f). The gray dashed line is DINOv2's own value, and the shaded bands denote $\pm1$ standard deviation. The two models are comparable at the image level, but \textcolor{gearred}{GEAR} is markedly closer to DINOv2 at the patch level (d,e) and in both locality statistics (c,f), showing that end-to-end guidance yields features with stronger local, spatially-causal structure.}
    \label{fig:repr_analysis}
\end{figure*}

To understand \emph{why} guiding the tokenizer with the AR model helps, we probe the tokenizer's own DINOv2 feature similarity (\cref{tab:tok_align}) and its codebook usage over training (\cref{fig:soft}), together with the DINOv2 similarity of the AR's per-layer hidden states (\cref{fig:repr_analysis}). Similarities are measured with CKNNA~\citep{cknna} and CKA~\citep{cka}. The emerging picture is the opposite of the diffusion-side recipe: end-to-end guidance leaves the tokenizer \emph{less} DINOv2-like, not more, and relocates the alignment into the AR.

\begin{table}[t]
    \centering
    \setlength{\tabcolsep}{5pt}
    \caption{\textbf{Tokenizer-feature alignment to DINOv2.} CKNNA and CKA between the tokenizer's own features and DINOv2 on the ImageNet validation set, before (pre) and after (post) quantization, at the image and patch level. Higher is more DINOv2-like. LlamaGen-REPA is the warm-up (non-end-to-end) tokenizer, and parenthesized values are GEAR's change relative to it.}
    \label{tab:tok_align}
    \begin{tabularx}{\linewidth}{>{\hsize=0.8\hsize\centering\arraybackslash}X >{\hsize=0.7\hsize\centering\arraybackslash}X|>{\hsize=0.9\hsize\centering\arraybackslash}X >{\hsize=1.35\hsize\centering\arraybackslash}X|>{\hsize=0.9\hsize\centering\arraybackslash}X >{\hsize=1.35\hsize\centering\arraybackslash}X}
        \toprule
        \multirow{2}{*}{\textbf{Metric}} & \multirow{2}{*}{\textbf{Level}} & \multicolumn{2}{c|}{\textbf{Pre-quantization}} & \multicolumn{2}{c}{\textbf{Post-quantization}} \\
        \cmidrule(lr){3-4} \cmidrule(lr){5-6}
         & & \makecell{LlamaGen-\\REPA} & GEAR & \makecell{LlamaGen-\\REPA} & GEAR \\
        \midrule
        \multirow{2}{*}{CKNNA} & image & 0.2485 & 0.2440~\textcolor{gray}{\scriptsize($-$0.0045)} & 0.2470 & 0.2410~\textcolor{gray}{\scriptsize($-$0.0059)} \\
         & patch & 0.1192 & 0.0831~\textcolor{red}{\small($-$0.0361)} & 0.1054 & 0.0748~\textcolor{red}{\small($-$0.0307)} \\
        \midrule
        \multirow{2}{*}{CKA}   & image & 0.1863 & 0.1805~\textcolor{gray}{\scriptsize($-$0.0058)} & 0.1846 & 0.1784~\textcolor{gray}{\scriptsize($-$0.0062)} \\
         & patch & 0.1727 & 0.1070~\textcolor{red}{\small($-$0.0657)} & 0.1609 & 0.0997~\textcolor{red}{\small($-$0.0612)} \\
        \bottomrule
    \end{tabularx}
\end{table}

\myparagraph{The tokenizer becomes less DINOv2-like.}
\cref{tab:tok_align} shows that end-to-end training makes the tokenizer's features \emph{less} similar to DINOv2, not more. GEAR is below the warm-up tokenizer at every entry, and the gap is far larger at the patch level (CKA $0.173$ to $0.107$) than at the image level. This is a re-organization rather than a loss of information, since reconstruction is preserved (\cref{tab:ablation_vq}). It is also harmless for generation, because the AR never consumes these continuous features. The AR ingests only the discrete index sequence through its own input embedding, so what it needs is not a DINOv2-like tokenizer but a discretization whose tokens it can predict and align with DINOv2. \textbf{Contrast this with latent diffusion.} REPA-E~\citep{repae}, VA-VAE~\citep{vavae} and MAETok~\citep{maetok} all improve diffusion by making the continuous VAE latent itself \emph{more} semantic and more aligned to vision-foundation features. Because a discrete AR consumes indices rather than a continuous latent, GEAR benefits from the reverse move. It reshapes the tokenizer for token predictability and lets the semantic alignment emerge inside the AR.

\myparagraph{The codebook usage sharpens.}
End-to-end training also reshapes \emph{which} codes the tokenizer uses, in the direction of predictability. \cref{fig:soft} tracks the codebook-assignment statistics over the joint fine-tuning run ($\tau{=}0.1$), starting from the warm-up tokenizer (blue). The AR rapidly sharpens the usage distribution: the cumulative top-$1$/$10$ assignment mass rises while the usage entropy and the effective codebook size fall, peaking in concentration around $30$k steps before relaxing and converging (after ${\sim}120$k) to a distribution that stays more concentrated and lower-entropy than the warm-up. A lower-entropy, more peaked token distribution is exactly an easier next-token target, so end-to-end training re-tunes the tokenizer to emit more predictable tokens. Because this pressure comes from the alignment signal rather than the prediction loss, it stops well short of the few-code collapse that next-token prediction would induce: the codebook stays broadly used and reconstruction is preserved (\cref{tab:ablation_vq}).

\myparagraph{Image-level vs. patch-level alignment.}
The AR's hidden states tell the opposite story. At the image level (a,b), the CKNNA and CKA curves of GEAR and LlamaGen-REPA are nearly indistinguishable across depth, both peaking around the alignment layer: explicit alignment already makes the two models comparably DINOv2-like in their global, pooled semantics. The picture changes sharply at the patch level (d,e): \textcolor{gearred}{GEAR (red)} lies well above \textcolor{reprablue}{LlamaGen-REPA (blue)} across the mid-to-deep layers, for \emph{both} CKNNA and CKA. In other words, the two methods agree with DINOv2 about the image as a whole, but GEAR matches DINOv2 far more closely \emph{per patch}, indicating that end-to-end guidance injects richer local structure into each token.

\myparagraph{Locality and spatial autocorrelation.}
Panels (c) and (f) corroborate this from a signal-processing view. Both Moran's~$I$ (c) and the FFT high-frequency ratio (f) measure how spatially self-correlated, or locally causal, the feature map is. GEAR tracks the DINOv2 reference (gray dashed) much more closely than LlamaGen-REPA: its Moran's~$I$ stays near the DINOv2 level at depth, and its high-frequency energy descends toward DINOv2's ratio rather than remaining elevated and noisy as for LlamaGen-REPA. Because autoregressive generation predicts each token causally from its local spatial context, tokens whose features carry coherent, DINOv2-like local structure are inherently easier to predict. This is the empirical signature of the AR model successfully reshaping the tokenizer's index distribution (\cref{fig:soft}), and it explains the consistent FID gains reported above.

\myparagraph{Where the alignment lives.}
Putting the two probes together, end-to-end training does not make the tokenizer more DINOv2-like. Instead it shifts the alignment burden from the tokenizer to the AR. The gradient routed back through the soft assignment reshapes the tokenizer toward a discretization the AR can predict and align, rather than toward DINOv2-like continuous features. As a result the tokenizer's own DINOv2 similarity drops (\cref{tab:tok_align}) while the AR's rises, most clearly per patch (\cref{fig:repr_analysis}). This is why a frozen end-to-end tokenizer trains a faster-converging and higher-quality AR (\cref{fig:trainloss,tab:comparison_detail}) even though, on its own, it looks \emph{less} semantic. Reading down the depth axis, the global metrics (a,b,c) peak near the alignment layer while the patch-level curves (d,e) keep strengthening into the deeper layers, and GEAR stays at least as close to DINOv2 as LlamaGen-REPA at \emph{every} depth. A detailed account of where each curve peaks, including the raw-feature high-frequency bump at $\ell{=}8$ in (f), is given in \cref{app:repr_peaks}.

\begin{table}[!t]
    \caption{\textbf{Ablation Study on Guidance Temperature.}}
    \centering
    \setlength{\tabcolsep}{1.2mm}{
    \begin{tabularx}{\linewidth}{A|*{5}{Y}|*{3}{Y}}
        \toprule
        \textbf{Temperature} & \textbf{gFID}$\downarrow$ & \textbf{sFID}$\downarrow$ & \textbf{IS}$\uparrow$ & \textbf{Prec.}$\uparrow$ & \textbf{Rec.}$\uparrow$ & \textbf{PSNR}$\uparrow$ & \textbf{SSIM}$\uparrow$ & \textbf{rFID}$\downarrow$ \\
         \midrule
        0.500 & \textbf{9.153} & 5.225 & \textbf{101.017} & \textbf{0.719} & 0.620 & 20.571 & 0.552 & 1.698 \\
        \rowcolor{blue!8} \textbf{0.100} & \underline{10.630} & 5.111 & 92.700 & \underline{0.701} & \underline{0.630} & 20.779 & 0.558 & 1.640 \\
        0.050 & 10.745 & \textbf{4.930} & \underline{93.024} & 0.696 & 0.626 & 20.804 & 0.558 & 1.638 \\
        0.010 & 11.424 & \underline{4.991} & 90.201 & 0.698 & 0.628 & 20.825 & 0.559 & 1.628 \\
        0.005 & 11.413 & 5.072 & 89.801 & 0.697 & \textbf{0.635} & 20.825 & 0.559 & 1.628 \\
        \bottomrule
    \end{tabularx}
    }
    \label{tab:diff_temperature}
\end{table}

\begin{table}[!t]
    \caption{\textbf{Ablation Study on Alignment Coefficient.}}
    \centering
    \setlength{\tabcolsep}{1.2mm}{
    \begin{tabularx}{\linewidth}{A|*{5}{Y}|*{3}{Y}}
        \toprule
        \textbf{Coefficient} & \textbf{gFID}$\downarrow$ & \textbf{sFID}$\downarrow$ & \textbf{IS}$\uparrow$ & \textbf{Prec.}$\uparrow$ & \textbf{Rec.}$\uparrow$ & \textbf{PSNR}$\uparrow$ & \textbf{SSIM}$\uparrow$ & \textbf{rFID}$\downarrow$ \\
         \midrule
        0.25 & 12.005 & \underline{5.165} & 87.183 & 0.691 & \textbf{0.637} & 20.912 & 0.564 & 1.653 \\
        \rowcolor{orange!8} \textbf{0.50} & \textbf{10.630} & \textbf{5.111} & \textbf{92.700} & \textbf{0.701} & 0.630 & 20.779 & 0.558 & 1.640 \\
        0.75 & \underline{10.819} & 5.426 & \underline{92.451} & \underline{0.696} & 0.620 & 20.618 & 0.551 & 1.647 \\
        1.00 & 10.889 & 5.507 & 91.086 & \underline{0.696} & \underline{0.633} & 20.465 & 0.546 & 1.715 \\
        \bottomrule
    \end{tabularx}
    }
    \label{tab:diff_coef}
\end{table}

\subsection{Ablation Studies}
\label{sec:ablation}

Each ablation table also reports tokenizer reconstruction: the warm-up tokenizer used to recover the GAN discriminator (the LlamaGen-REPA row) and GEAR's end-to-end-tuned tokenizer (the GEAR row). Across all studies, end-to-end training preserves and often slightly improves reconstruction. \cref{app:tok_recon} details the official tokenizers and how the evaluation interpolation affects these reconstruction metrics.

\abpar{blue!8}{Guidance temperature.}
\cref{tab:diff_temperature} sweeps the temperature $\tau$ of the soft assignment. As $\tau$ shrinks, the soft batch collapses onto the hard one and the guidance signal weakens, so gFID degrades monotonically below $\tau{=}0.1$ while rFID improves only marginally. A large $\tau{=}0.5$ yields the lowest gFID ($9.153$) but the loosest soft-hard consistency and the worst reconstruction ($1.698$ rFID). We adopt $\tau{=}0.1$ as a robust default that balances generation and reconstruction across the other studies.

\abpar{orange!8}{Alignment coefficient.}
\cref{tab:diff_coef} varies the coefficient $\lambda$ that weights both alignment terms. Performance is best at $\lambda{=}0.5$ on essentially every metric: a smaller $\lambda{=}0.25$ under-uses the guidance, while a larger $\lambda{=}1.0$ over-regularizes and hurts both generation and reconstruction (rFID $1.715$). We therefore set $\lambda{=}0.5$.

\abpar{cyan!8}{Role of different components.}
\cref{tab:ablation} isolates two design choices. Replacing our differentiable soft-assignment bridge with the conventional straight-through estimator destabilizes the joint training catastrophically, collapsing both generation (gFID $104.9$) and reconstruction (rFID $59.7$). This confirms that the soft assignment is essential as the differentiable channel that carries AR gradients into the tokenizer. Removing the adversarial loss is far less severe but still markedly degrades both reconstruction ($1.640{\to}5.857$ rFID) and generation ($10.630{\to}16.353$ gFID), confirming that adversarial supervision remains important for the co-trained tokenizer.

\abpar{OliveGreen!8}{VQ tokenizers.}
\cref{tab:ablation_vq} shows that GEAR is largely agnostic to the underlying quantizer, improving all three: on VQVAE~\citep{vqvae,llamagen} gFID drops $14.72\!\to\!10.63$ (rFID $1.724\!\to\!1.640$), on LFQ~\citep{magvitv2,openmagvitv2} $18.68\!\to\!14.78$ (rFID $2.421\!\to\!2.129$), and on IBQ~\citep{ibq} $20.25\!\to\!12.97$ (rFID $1.973\!\to\!1.716$), the largest gain. In every case GEAR improves both generation and reconstruction, confirming that the guidance mechanism is independent of the quantization scheme.

\abpar{red!8}{Model size.}
\cref{tab:different_size} scales GEAR on top of LlamaGen-REPA at the B, L and XL sizes. As expected, generation improves monotonically with model size (gFID $21.52\!\to\!10.63\!\to\!7.69$, IS $54.4\!\to\!92.7\!\to\!115.5$). More surprisingly, the reconstruction quality of the \emph{co-trained} tokenizer also improves monotonically (PSNR $20.751\!\to\!20.808$, SSIM $0.556\!\to\!0.559$ and rFID $1.658\!\to\!1.624$ from B to XL), whereas the frozen baseline tokenizer is by construction identical across sizes (rFID $1.724$). Because the tokenizer is supervised only indirectly, through the soft guidance, this indicates that a stronger AR model provides a better guidance signal, yielding a tokenizer that is at once easier to sample from and higher in fidelity.

\begin{table}[!t]
    \caption{\textbf{Ablation Study on Role of Different Components.}}
    \centering
    \setlength{\tabcolsep}{1.2mm}{
    \begin{tabularx}{\linewidth}{A|*{5}{Y}|*{3}{Y}}
        \toprule
        \textbf{Component} & \textbf{gFID}$\downarrow$ & \textbf{sFID}$\downarrow$ & \textbf{IS}$\uparrow$ & \textbf{Prec.}$\uparrow$ & \textbf{Rec.}$\uparrow$ & \textbf{PSNR}$\uparrow$ & \textbf{SSIM}$\uparrow$ & \textbf{rFID}$\downarrow$ \\
         \midrule
         w/ STE & 104.932 & 31.933 & 12.418 & 0.292 & 0.233 & 12.596 & 0.239 & 59.723 \\
         w/o $\mathcal{L}_{\mathrm{GAN}}$ & 16.353 & 8.084 & 76.346 & 0.649 & 0.590 & 22.343 & 0.606 & 5.857 \\
        \rowcolor{cyan!8} \textbf{GEAR (Ours)} & \textbf{10.630} & \textbf{5.111} & \textbf{92.700} & \textbf{0.701} & \textbf{0.630} & 20.779 & 0.558 & 1.640 \\
        \bottomrule
    \end{tabularx}
    }
    \label{tab:ablation}
\end{table}

\begin{table}[!t]
    \caption{\textbf{Ablation Study on VQ Tokenizers.}}
    \centering
    \setlength{\tabcolsep}{1.2mm}{
    \begin{tabularx}{\linewidth}{A|*{5}{Y}|*{3}{Y}}
        \toprule
        \textbf{Model} & \textbf{gFID}$\downarrow$ & \textbf{sFID}$\downarrow$ & \textbf{IS}$\uparrow$ & \textbf{Prec.}$\uparrow$ & \textbf{Rec.}$\uparrow$ & \textbf{PSNR}$\uparrow$ & \textbf{SSIM}$\uparrow$ & \textbf{rFID}$\downarrow$ \\
        \midrule
        VQVAE~\cite{vqvae,llamagen} & 14.719 & 6.009 & 77.333 & 0.670 & \textbf{0.640} & 21.061 & 0.565 & 1.724 \\
        \rowcolor{OliveGreen!8} \textbf{+GEAR (Ours)} & \textbf{10.630} & \textbf{5.111} & \textbf{92.700} & \textbf{0.701} & 0.630 & 20.779 & 0.558 & 1.640 \\
        \midrule
        LFQ~\cite{magvitv2,openmagvitv2} & 18.681 & 6.525 & 65.265 & 0.666 & \textbf{0.625} & 20.969 & 0.562 & 2.421 \\
        \rowcolor{OliveGreen!8} \textbf{+GEAR (Ours)} & \textbf{14.776} & \textbf{5.553} & \textbf{74.512} & \textbf{0.690} & 0.621 & 20.477 & 0.550 & 2.129 \\
        \midrule
        IBQ~\cite{ibq} & 20.246 & 6.897 & 61.628 & 0.638 & \textbf{0.634} & 21.182 & 0.577 & 1.973 \\
        \rowcolor{OliveGreen!8} \textbf{+GEAR (Ours)} & \textbf{12.972} & \textbf{5.165} & \textbf{80.307} & \textbf{0.679} & 0.629 & 20.917 & 0.567 & 1.716 \\
        \bottomrule
    \end{tabularx}
    }
    
    \label{tab:ablation_vq}
\end{table}

\begin{table}[!t]
    \caption{\textbf{Ablation Study on Model Size.}}
    \centering
    \setlength{\tabcolsep}{1.2mm}{
    \begin{tabularx}{\linewidth}{A|*{5}{Y}|*{3}{Y}}
        \toprule
        \textbf{Model} & \textbf{gFID}$\downarrow$ & \textbf{sFID}$\downarrow$ & \textbf{IS}$\uparrow$ & \textbf{Prec.}$\uparrow$ & \textbf{Rec.}$\uparrow$ & \textbf{PSNR}$\uparrow$ & \textbf{SSIM}$\uparrow$ & \textbf{rFID}$\downarrow$ \\
        \midrule
        LlamaGen-REPA-B~\cite{llamagen-repa} & 24.986 & 6.762 & 49.086 & 0.620 & \textbf{0.614} & 21.061 & 0.565 & 1.724 \\
        \rowcolor{red!8} \textbf{+GEAR (Ours)} & \textbf{21.516} & \textbf{5.420} & \textbf{54.397} & \textbf{0.644} & 0.610 & 20.751 & 0.556 & 1.658 \\
        \midrule
        LlamaGen-REPA-L~\cite{llamagen-repa} & 14.719 & 6.009 & 77.333 & 0.670 & \textbf{0.640} & 21.061 & 0.565 & 1.724 \\
        \rowcolor{red!8} \textbf{+GEAR (Ours)} & \textbf{10.630} & \textbf{5.111} & \textbf{92.700} & \textbf{0.701} & 0.630 & 20.779 & 0.558 & 1.640 \\
        \midrule
        LlamaGen-REPA-XL~\cite{llamagen-repa} & 9.631 & 5.556 & 104.456 & 0.697 & \textbf{0.644} &  21.061 & 0.565 & 1.724 \\
        \rowcolor{red!8} \textbf{+GEAR (Ours)} & \textbf{7.693} & \textbf{4.934} & \textbf{115.463} & \textbf{0.717} & 0.637 & 20.808 & 0.559 & 1.624 \\
        \bottomrule
    \end{tabularx}
    }
    \label{tab:different_size}
\end{table}

\begin{table}[!t]
    \caption{\textbf{Ablation Study on Representation Encoder.} The baselines and GEAR are both trained for $100$k steps.}
    \centering
    \setlength{\tabcolsep}{1.2mm}{
    \begin{tabularx}{\linewidth}{A|*{5}{Y}|*{3}{Y}}
        \toprule
        \textbf{Model} & \textbf{gFID}$\downarrow$ & \textbf{sFID}$\downarrow$ & \textbf{IS}$\uparrow$ & \textbf{Prec.}$\uparrow$ & \textbf{Rec.}$\uparrow$ & \textbf{PSNR}$\uparrow$ & \textbf{SSIM}$\uparrow$ & \textbf{rFID}$\downarrow$ \\
        \midrule
        DINOv2-B~\cite{dinov2} & 22.371 & 6.522 & 52.982 & 0.635 & \textbf{0.620} & 21.061 & 0.565 & 1.724  \\
        \rowcolor{yellow!8} \textbf{+GEAR (Ours)} & \textbf{16.837} & \textbf{5.408} & \textbf{65.098} & \textbf{0.670} & 0.611 & 20.925 & 0.561 & 1.721 \\
        \midrule
        DINOv3-B~\cite{dinov3} & 23.115 & 6.425 & 51.483 & 0.631 & \textbf{0.621} & 21.061 & 0.565 & 1.724 \\
        \rowcolor{yellow!8} \textbf{+GEAR (Ours)} & \textbf{18.967} & \textbf{5.597} & \textbf{59.838} & \textbf{0.658} & \textbf{0.621} & 21.021 & 0.566 & 1.696 \\
        \midrule
        SigLIPv2-B~\cite{siglip2} & 23.254 & 6.369 & 50.830 & 0.631 & \textbf{0.627} & 21.061 & 0.565 & 1.724 \\
        \rowcolor{yellow!8} \textbf{+GEAR (Ours)} & \textbf{19.226} & \textbf{5.360} & \textbf{59.047} & \textbf{0.656} & 0.611 & 20.867 & 0.558 & 1.803 \\
        \midrule
        V-JEPA2.1-B~\cite{vjepa21} & 23.536 & 6.369 & 50.216 & 0.638 & 0.608 & 21.061 & 0.565 & 1.724  \\
        \rowcolor{yellow!8} \textbf{+GEAR (Ours)} & \textbf{19.532} & \textbf{5.876} & \textbf{56.754} & \textbf{0.653} & \textbf{0.610} & 20.759 & 0.553 & 1.879 \\
        \bottomrule
    \end{tabularx}
    }
    
    \label{tab:different_repr}
\end{table}

\begin{table}[!t]
    \caption{\textbf{Ablation Study on Alignment Depth.} The baselines and GEAR are both trained for $100$k steps.}
    \centering
    \setlength{\tabcolsep}{1.2mm}{
    \begin{tabularx}{\linewidth}{A|*{5}{Y}|*{3}{Y}}
        \toprule
        \textbf{Model} & \textbf{gFID}$\downarrow$ & \textbf{sFID}$\downarrow$ & \textbf{IS}$\uparrow$ & \textbf{Prec.}$\uparrow$ & \textbf{Rec.}$\uparrow$ & \textbf{PSNR}$\uparrow$ & \textbf{SSIM}$\uparrow$ & \textbf{rFID}$\downarrow$ \\
         \midrule
        6th layer & 23.102 & 6.428 & 51.912 & 0.636 & \textbf{0.618} & 21.061 & 0.565 & 1.724 \\
        \rowcolor{green!8} \textbf{+GEAR (Ours)} & \textbf{16.961} & \textbf{5.331} & \textbf{64.200} & \textbf{0.673} & 0.609 & 20.889 & 0.560 & 1.714 \\
         \midrule
        8th layer & 22.371 & 6.522 & 52.982 & 0.635 & \textbf{0.620} & 21.061 & 0.565 & 1.724 \\
        \rowcolor{green!8} \textbf{+GEAR (Ours)} & \textbf{16.837} & \textbf{5.408} & \textbf{65.098} & \textbf{0.670} & 0.611 & 20.925 & 0.561 & 1.721 \\
         \midrule
        10th layer & 22.455 & 6.417 & 52.299 & 0.633 & \textbf{0.628} & 21.061 & 0.565 & 1.724 \\
        \rowcolor{green!8} \textbf{+GEAR (Ours)} & \textbf{17.988} & \textbf{5.380} & \textbf{61.915} & \textbf{0.666} & 0.616 & 20.930 & 0.561 & 1.714 \\
        \bottomrule
    \end{tabularx}
    }
    
    \label{tab:diff_depth}
\end{table}

\begin{table}[!t]
    \caption{\textbf{Ablation Study on Tokenizer Initialization.} ``w/ init.''\ starts the GEAR-L tokenizer from the warm-up checkpoint (our default), while ``w/o init.''\ trains it from scratch jointly with the AR.}
    \centering
    \setlength{\tabcolsep}{1.2mm}{
    \begin{tabularx}{\linewidth}{A|*{5}{Y}|*{3}{Y}}
        \toprule
        \textbf{Model} & \textbf{gFID}$\downarrow$ & \textbf{sFID}$\downarrow$ & \textbf{IS}$\uparrow$ & \textbf{Prec.}$\uparrow$ & \textbf{Rec.}$\uparrow$ & \textbf{PSNR}$\uparrow$ & \textbf{SSIM}$\uparrow$ & \textbf{rFID}$\downarrow$ \\
        \midrule
        LlamaGen-REPA-L~\cite{llamagen-repa} & 14.719 & 6.009 & 77.333 & 0.670 & \textbf{0.640} & \textbf{21.061} & \textbf{0.565} & 1.724 \\
        \midrule
        GEAR (w/o init.) & 13.435 & 5.810 & 80.696 & 0.689 & 0.619 & 20.515 & 0.547 & 2.256 \\
        \rowcolor{purple!8} \textbf{GEAR (w/ init.)} & \textbf{10.630} & \textbf{5.111} & \textbf{92.700} & \textbf{0.701} & 0.630 & 20.779 & 0.558 & \textbf{1.640} \\
        \bottomrule
    \end{tabularx}
    }
    \label{tab:diff_init}
\end{table}

\abpar{yellow!8}{Representation encoder.}
\cref{tab:different_repr} replaces the alignment target. GEAR improves generation regardless of the choice of pretrained encoder, lowering gFID by $4$--$6$ points for DINOv2~\citep{dinov2}, DINOv3~\citep{dinov3}, SigLIPv2~\citep{siglip2} and V-JEPA2.1~\citep{vjepa21}. DINOv2 gives the strongest result ($16.837$ gFID) and is used as our default target.

\abpar{green!8}{Alignment depth.}
\cref{tab:diff_depth} varies the AR layer at which alignment is applied. GEAR improves over the corresponding baseline at every depth, and the $8$-th layer attains the best gFID ($16.837$), which we adopt by default.

\abpar{purple!8}{Tokenizer initialization.}
\cref{tab:diff_init} asks whether GEAR needs a pretrained tokenizer to start from. Initializing the end-to-end tokenizer from the warm-up checkpoint (our default) is best (gFID $10.630$, rFID $1.640$). Training the tokenizer \emph{from scratch} jointly with the AR still beats the frozen pretrained baseline on generation (gFID $13.435$ vs.\ $14.719$ for LlamaGen-REPA), though its reconstruction is worse (rFID $2.256$). Initialization is thus not required for the guidance to help, but it gives a sizable extra boost and keeps the tokenizer's reconstruction intact, so we initialize from the warm-up tokenizer by default.

\abpar{Mulberry!8}{Resolution.}
\cref{tab:diff_resolution} extends GEAR to higher generation resolutions ($384$ and $512$). Following LlamaGen, images are sampled at the training resolution and resized to $256$ before computing FID. GEAR improves generation over LlamaGen-REPA at every resolution and under both sampling settings: with CFG it lowers the gFID from $7.26$ to $5.49$ at $256$, $7.16$ to $5.95$ at $384$, and $10.20$ to $7.60$ at $512$ (with consistently higher IS), confirming that the end-to-end guidance transfers to higher resolutions.

\begin{table}[!t]
    \caption{\textbf{Ablation Study on Resolution.} All models are trained for $100$k steps. Following LlamaGen~\cite{llamagen}, we sample at the training resolution (e.g., $384$, $512$) and resize to $256$ before computing FID. The w/ CFG columns use a guidance scale of $1.5$.}
    \centering
    \setlength{\tabcolsep}{1.2mm}{
    \begin{tabularx}{\linewidth}{l|*{5}{Y}|*{5}{Y}}
        \toprule
        \multirow{2}{*}{\textbf{Resolution}} & \multicolumn{5}{c|}{\textbf{Generation w/o CFG}} & \multicolumn{5}{c}{\textbf{Generation w/ CFG}} \\
        \cmidrule(lr){2-6} \cmidrule(lr){7-11}
        & \textbf{gFID}$\downarrow$ & \textbf{sFID}$\downarrow$ & \textbf{IS}$\uparrow$ & \textbf{Prec.}$\uparrow$ & \textbf{Rec.}$\uparrow$ & \textbf{gFID}$\downarrow$ & \textbf{sFID}$\downarrow$ & \textbf{IS}$\uparrow$ & \textbf{Prec.}$\uparrow$ & \textbf{Rec.}$\uparrow$ \\
         \midrule
        256 & 22.371 & 6.522 & 52.982 & 0.635 & \textbf{0.620} & 7.255 & 5.587 & 127.827 & 0.788 & \textbf{0.531} \\
        \rowcolor{Mulberry!8} \textbf{+GEAR (Ours)} & \textbf{16.837} & \textbf{5.408} & \textbf{65.098} & \textbf{0.670} & 0.611 & \textbf{5.492} & \textbf{5.167} & \textbf{157.681} & \textbf{0.822} & 0.511 \\
        \midrule
        384 & 22.977 & 7.226 & 54.079 & 0.616 & \textbf{0.631} & 7.159 & 5.739 & 130.327 & 0.772 & \textbf{0.554} \\
        \rowcolor{Mulberry!8} \textbf{+GEAR (Ours)} & \textbf{18.364} & \textbf{5.874} & \textbf{62.428} & \textbf{0.658} & 0.605 & \textbf{5.945} & \textbf{5.383} & \textbf{143.497} & \textbf{0.807} & 0.508 \\
        \midrule
        512 & 27.742 & 8.720 & 46.125 & 0.587 & \textbf{0.633} & 10.204 & 6.650 & 107.186 & 0.736 & \textbf{0.553} \\
        \rowcolor{Mulberry!8} \textbf{+GEAR (Ours)} & \textbf{21.826} & \textbf{6.336} & \textbf{55.689} & \textbf{0.634} & 0.613 & \textbf{7.595} & \textbf{5.621} & \textbf{126.386} & \textbf{0.779} & 0.531 \\
        \bottomrule
    \end{tabularx}
    }
    \label{tab:diff_resolution}
\end{table}

\begin{wraptable}{r}{0.5\textwidth}
    \vspace{-1.25em}
    \caption{\textbf{Ablation Study on CFG Scale.}}
    \vspace{-0.5em}
    \centering
    \setlength{\tabcolsep}{1.2mm}
    \small
    \begin{tabular}{l|ccccc}
        \toprule
        \textbf{CFG Scale} & \textbf{gFID}$\downarrow$ & \textbf{sFID}$\downarrow$ & \textbf{IS}$\uparrow$ & \textbf{Prec.}$\uparrow$ & \textbf{Rec.}$\uparrow$ \\
        \midrule
        1.00 (w/o CFG) & 10.630 & \textbf{5.111} & 92.700 & 0.701 & \textbf{0.630} \\
        1.25 & 4.811 & 5.114 & 154.546 & 0.786 & 0.570 \\
        \rowcolor{NavyBlue!8}
        \textbf{1.50} & \textbf{3.388} & 5.446 & 216.915 & 0.840 & 0.525 \\
        1.75 & 3.916 & 5.928 & 268.230 & 0.872 & 0.471 \\
        2.00 & 5.264 & 6.540 & \textbf{309.665} & \textbf{0.895} & 0.431 \\
        \bottomrule
    \end{tabular}
    \label{tab:cfg_sweep}
    \vspace{-1em}
\end{wraptable}

\abpar{NavyBlue!8}{Classifier-free guidance.} \cref{tab:cfg_sweep} sweeps the CFG scale to find the best trade-off between sample quality and diversity. Generation follows the familiar pattern: increasing the scale rapidly improves gFID and IS while reducing Recall, with the best gFID of $3.388$ reached at a scale of $1.5$ (from $10.63$ without guidance). We therefore adopt a CFG scale of $1.5$ for guided sampling.

\section{Discussion}
\label{sec:discussion}

\noindent\textbf{Reconstruction ceiling.}
Under a far smaller training budget, GEAR substantially narrows the gap between autoregressive generation and strong latent-diffusion baselines, yet it still trails the best end-to-end diffusion model, REPA-E~\cite{repae}. The reason is the discrete tokenizer. GEAR's reconstruction (rFID $1.64$) upper-bounds its generation (gFID $2.52$ with CFG), whereas REPA-E's continuous VAE reconstructs far more faithfully (rFID $0.28$) and reaches gFID $1.12$. Closing this reconstruction gap is the single largest lever for further improving VQ-AR generation.

\noindent\textbf{Compression is coupled to compute in autoregression.}
The deeper cause is architectural. In current VQ-AR pipelines the tokenizer down-samples by $16\times$, mapping a $256\times256$ image to $256$ tokens, and the AR model spends its compute on exactly those $256$ tokens, so the compression rate and the sequence length are tied together. Latent diffusion instead decouples them. It uses a milder $8\times$ tokenizer that keeps $1024$ latent positions, and thus higher fidelity, then applies a $2{\times}2$ patch embedding so the transformer still operates on $256$ tokens. Because one token is one decoding step, an AR model cannot lengthen its latent without lengthening its sequence and its compute, and is therefore pushed toward a more aggressive, lossier tokenizer. Borrowing this decoupling from the diffusion side, for instance a milder tokenizer paired with AR-side grouping such as patchified or multi-token prediction, is a promising route to raise the reconstruction ceiling without inflating the sequence.

\noindent\textbf{Toward unified, long-context generation.}
Despite this ceiling, the next-token formulation remains attractive for several reasons. It is uniform across modalities and friendly to scaling and engineering. Its discrete tokens also bound the per-step error, whereas continuous latents accumulate it over a long context, and unified pipelines that re-encode and decode across heterogeneous understanding and generation encoders, such as BAGEL~\cite{bagel}, compound this drift across turns. Finally, the discrete next-token form directly inherits the mature alignment stack of large language models, from reinforcement learning from human feedback~\cite{rlhf} to preference-optimization methods such as PPO~\cite{ppo}, DPO~\cite{dpo} and GRPO~\cite{grpo}. Such preference alignment is almost always the final and indispensable step before a model is deployed in practice. End-to-end discrete VQ-AR, as instantiated by GEAR, is therefore a promising substrate for unified, long-context understanding and generation.

\section{Conclusion}
\label{sec:conclusion}

We presented GEAR, a framework that trains a VQ tokenizer and an autoregressive generator jointly and end-to-end. The key idea is to let the AR model's representation-alignment objective guide the tokenizer through a differentiable soft assignment, while a hard, one-hot branch trains the generator on exactly the discrete tokens used at inference. Routing only the alignment signal, and never the prediction loss, into the tokenizer overcomes the non-differentiable index that defeats the straight-through estimator and avoids the code collapse it would otherwise induce. Across class-conditional ImageNet and text-to-image generation, GEAR substantially accelerates training relative to the strong LlamaGen-REPA baseline (up to $10\times$ faster ImageNet gFID convergence) and improves final quality, while also improving the reconstruction of the co-trained tokenizer. A representation analysis shows that these gains come from sharper patch-level, spatially-coherent structure, which is exactly what makes next-token prediction easier, and the mechanism is a drop-in across quantizers (VQVAE, LFQ, IBQ). We believe guided end-to-end training is a general principle for visual generation, and a promising next step is to scale it to larger text-to-image models and to unified understanding-and-generation systems.

\clearpage

\bibliographystyle{plainnat}
\setlength{\bibhang}{0pt}
\setlength\bibindent{0pt}
\bibliography{main}

\clearpage
\appendix
\section{Ablation Training Configurations}
\label{app:ablation_configs}

All ablation studies fine-tune the GEAR-L model end-to-end on top of the warm-up tokenizer (except the initialization study, which also trains the tokenizer from scratch), and each study varies a single axis while holding everything else fixed. \cref{tab:ablation_configs} gives one column per study, with the axis each study sweeps shown in \textbf{bold} (default \underline{underlined}). The optimization, loss-weight and sampling settings shared by all studies are listed separately in \cref{tab:ablation_shared}.

A few conventions deserve a note. The alignment target is spatially normalized following iREPA~\citep{irepa} (per-channel $z$-score with strength $\alpha{=}0.6$) only for the SigLIPv2 and V-JEPA2.1 targets, which need it to serve as good REPA targets. DINOv2 and DINOv3 use no spatial normalization, as in the original REPA~\citep{repa}. Across model sizes the alignment is always applied at one-third of the depth: the B/L/XL backbones have $12/24/36$ layers, so the alignment layer is $4/8/12$ respectively, which is the optimal alignment depth identified by LlamaGen-REPA~\citep{llamagen-repa}. The reconstruction loss weights are read off the tokenizer objective of \cref{eq:vqloss}.

\begin{table}[h]
    \centering
    \footnotesize
    \setlength{\tabcolsep}{4pt}
    \caption{\textbf{Per-study training configurations for the ablation studies (Tables~\ref{tab:diff_temperature}--\ref{tab:diff_resolution}).} Each column is one study and lists its configuration, with the axis it sweeps in \textbf{bold} (default \underline{underlined}). Settings shared by all studies are listed in \cref{tab:ablation_shared}.}
    \label{tab:ablation_configs}
    \renewcommand{\tabularxcolumn}[1]{m{#1}}%
    \begin{tabularx}{\linewidth}{@{}>{\centering\arraybackslash}m{2.7cm}|*{9}{Y}@{}}
        \toprule
        \textbf{Setting}
            & \makecell{Temp.\\{\scriptsize Tab.~\ref{tab:diff_temperature}}}
            & \makecell{Coeff.\\{\scriptsize Tab.~\ref{tab:diff_coef}}}
            & \makecell{Comp.\\{\scriptsize Tab.~\ref{tab:ablation}}}
            & \makecell{Tok.\\{\scriptsize Tab.~\ref{tab:ablation_vq}}}
            & \makecell{Size\\{\scriptsize Tab.~\ref{tab:different_size}}}
            & \makecell{Enc.\\{\scriptsize Tab.~\ref{tab:different_repr}}}
            & \makecell{Depth\\{\scriptsize Tab.~\ref{tab:diff_depth}}}
            & \makecell{Init\\{\scriptsize Tab.~\ref{tab:diff_init}}}
            & \makecell{Res.\\{\scriptsize Tab.~\ref{tab:diff_resolution}}} \\
        \midrule
        AR model & L & L & L & L & \makecell{\textbf{B}\\\textbf{\underline{L}}\\\textbf{XL}} & L & L & L & L \\
        \midrule
        Alignment layer $\ell$ & 8 & 8 & 8 & 8 & \makecell{4\\\underline{8}\\12} & 8 & \makecell{\textbf{6}\\\textbf{\underline{8}}\\\textbf{10}} & 8 & 8 \\
        \midrule
        REPA target & DINOv2 & DINOv2 & DINOv2 & DINOv2 & DINOv2 & \makecell{\textbf{\underline{DINOv2}}\\\textbf{DINOv3}\\\textbf{SigLIPv2}\\\textbf{V\text{-}JEPA2.1}} & DINOv2 & DINOv2 & DINOv2 \\
        \midrule
        Tokenizer & VQ-16 & VQ-16 & VQ-16 & \makecell{\textbf{\underline{VQ-16}}\\\textbf{LFQ-16}\\\textbf{IBQ-16}} & VQ-16 & VQ-16 & VQ-16 & VQ-16 & VQ-16 \\
        \midrule
        Gradient & soft & soft & \makecell{\textbf{\underline{soft}}\\\textbf{STE}\\\textbf{w/o $\mathcal{L}_{\mathrm{GAN}}$}} & soft & soft & soft & soft & soft & soft \\
        \midrule
        Guidance temp.\ $\tau$ & \makecell{\textbf{0.5}\\\textbf{\underline{0.1}}\\\textbf{0.05}\\\textbf{0.01}\\\textbf{0.005}} & 0.1 & 0.1 & 0.1 & 0.1 & 0.1 & 0.1 & 0.1 & 0.1 \\
        \midrule
        Align.\ coeff.\ $\lambda$ & 0.5 & \makecell{\textbf{0.25}\\\textbf{\underline{0.5}}\\\textbf{0.75}\\\textbf{1.0}} & 0.5 & 0.5 & 0.5 & 0.5 & 0.5 & 0.5 & 0.5 \\
        \midrule
        Resolution & 256 & 256 & 256 & 256 & 256 & 256 & 256 & 256 & \makecell{\textbf{\underline{256}}\\\textbf{384}\\\textbf{512}} \\
        \midrule
        Training steps & 400k & 400k & 400k & 400k & 400k & 100k & 100k & 400k & 100k \\
        \midrule
        Tokenizer init. & \cmark & \cmark & \cmark & \cmark & \cmark & \cmark & \cmark & \textbf{\xmark} & \cmark \\
        \bottomrule
    \end{tabularx}
\end{table}

\begin{table}[t]
    \centering
    \setlength{\tabcolsep}{4pt}
    \caption{\textbf{Shared training configuration (Tables~\ref{tab:diff_temperature}--\ref{tab:diff_resolution} and Tables~\ref{tab:comparison_detail}--\ref{tab:gpic_t2i_performance}).} Optimization, loss-weight and sampling settings used by every ablation study.}
    \label{tab:ablation_shared}
    \begin{tabularx}{\linewidth}{@{}>{\raggedright\arraybackslash}p{4.0cm} Y >{\raggedright\arraybackslash}p{4.0cm} Y@{}}
        \toprule
        Batch size & 256 & Precision & bf16 \\
        Learning rate & $1{\times}10^{-4}$ & Codebook size & 16384 \\
        Optimizer & AdamW & CLS dropout & 0.1 \\
        Adam $\beta_1$ & 0.9 & Projector dim & 2048 \\
        Adam $\beta_2$ & 0.999 & EMA decay & 0.9999 \\
        Weight decay & 0 & torch.compile & \cmark \\
        Gradient clip & 1.0 & Entropy & 0.05 \\
        NTP loss & 1.0 & Commitment & 0.25 \\
        Reconstruction (L1) & 1.0 & Sampling temperature & 1.0 \\
        Perceptual (LPIPS) & 0.1 & Top-$k$ & 0 \\
        Adversarial (GAN) & 0.1 & Top-$p$ & 1.0 \\
        \bottomrule
    \end{tabularx}
\end{table}

\section{Text-to-Image Training Configurations}
\label{app:t2i}

Our text-to-image model is a strict autoregressor over the concatenation of the text and image tokens. Following an MMDiT-style hybrid design, the first third of the transformer blocks are \emph{dual-stream} (separate query/key/value/output projections for the text and image tokens) and the remaining two thirds are \emph{single-stream} with shared projections, while the attention stays purely causal throughout. The text condition is a $300$-token Qwen3-1.7B embedding with a $0.1$ caption-dropping probability for classifier-free guidance, and a DINOv2 REPA loss is applied as in the class-conditional setting.

The model is trained \textbf{from scratch} on a LlamaGen-1B backbone, which matches the parameter count of the GPIC JiT-1B baseline we compare against, for a single epoch over the $\sim$$100$M-image GPIC corpus at $256$ resolution (about $390$k steps at batch size $256$) with a constant learning rate. Since there is no class-conditional checkpoint to inherit, the text stream is learned from scratch as well. GEAR and LlamaGen-REPA follow this identical recipe and differ only in whether the frozen tokenizer has been end-to-end fine-tuned. \cref{tab:gpic_train_config} lists the configuration.

\begin{table}[t]
    \centering
    \setlength{\tabcolsep}{4pt}
    \caption{\textbf{Text-to-image training configuration on GPIC.} GEAR and LlamaGen-REPA share this recipe and differ only in whether the frozen tokenizer has been end-to-end fine-tuned. The model is trained from scratch, and the LlamaGen-1B backbone matches the parameter count of the GPIC JiT-1B baseline.}
    \label{tab:gpic_train_config}
    \begin{tabularx}{\linewidth}{@{}>{\raggedright\arraybackslash}p{4.0cm} Y >{\raggedright\arraybackslash}p{3.0cm} Y@{}}
        \toprule
        \multicolumn{4}{@{}l}{\textbf{Frozen VQ tokenizer.}\quad \textbf{GEAR}: warm-up, then end-to-end fine-tuned.\quad \textbf{LlamaGen-REPA}: warm-up only.} \\
        \midrule\midrule
        AR model & LlamaGen-1B & AR initialization & \textbf{from scratch} \\
        Align layer & 12 & Resolution & 256 \\
        Training data & GPIC ($\sim$100M) & Schedule & 1 epoch ($\sim$390k) \\
        LR schedule & constant & AR learning rate & $1{\times}10^{-4}$ \\
        Batch size & 256 & Text encoder & Qwen3-1.7B \\
        Text length & 300 & Caption dropout & 0.1 \\
        Attention & causal & REPA target & DINOv2 \\
        Align.\ coeff.\ $\lambda$ & 0.5 & Optimizer & AdamW (0.9, 0.999) \\
        Weight decay & 0 & Gradient clip & 1.0 \\
        Precision & bf16 & Codebook size & 16384 \\
        EMA decay & 0.9999 & torch.compile & \cmark \\
        \midrule\midrule
        Sampling temperature & 1.0 & Top-$k$ & 0 \\
        Top-$p$ & 1.0 & & \\
        \bottomrule
    \end{tabularx}
\end{table}

\section{Tokenizer Reconstruction and Evaluation Interpolation}
\label{app:tok_recon}

We instantiate GEAR on three publicly released tokenizers, all with a $16384$-entry codebook: VQ-16 from LlamaGen~\citep{llamagen}, LFQ-16 from Open-MAGVIT2~\citep{openmagvitv2}, and IBQ-16~\citep{ibq}. \cref{tab:tok_recon} reports their sizes and reconstruction quality on the ImageNet validation set.

One pipeline detail materially affects these numbers: the interpolation used when a validation image is short-side-resized and center-cropped to the evaluation resolution before being fed to the tokenizer. Since PSNR, SSIM and rFID are all computed against this resized image, the interpolation changes the reference itself. Bicubic keeps more high-frequency content (sharper) while bilinear is smoother, so the two give different scores. We evaluate with \textbf{bicubic}, whereas the official LFQ and IBQ numbers are reported with bilinear. Re-evaluating the released weights with bilinear reproduces the official numbers closely (for example, LFQ rFID $2.55$ versus our $2.56$), confirming that the gap reflects the interpolation rather than a different checkpoint. The \emph{warm-up} and \emph{GEAR} rows repeat the reconstruction metrics from \cref{tab:ablation_vq} (also measured with bicubic), showing that recovering the GAN discriminator and then training end-to-end preserves reconstruction. The official LlamaGen SSIM (marked $^{*}$) is inflated because its implementation computes SSIM treating $[0,1]$-valued images as $[-1,1]$.

\begin{table}[t]
    \centering
    \footnotesize
    \setlength{\tabcolsep}{4pt}
    \caption{\textbf{Tokenizer reconstruction on the ImageNet validation set.} \emph{Setting} is the source of each row: \emph{Reported} from the official paper, \emph{Reproduced} our re-evaluation of the released weights, \emph{Warm-up} after recovering the GAN discriminator, and \emph{GEAR} after end-to-end training (the last two repeat \cref{tab:ablation_vq}). We evaluate with bicubic, whereas the official LFQ/IBQ numbers use bilinear. $^{*}$LlamaGen's official SSIM is computed treating $[0,1]$ images as $[-1,1]$ and is therefore inflated.}
    \label{tab:tok_recon}
    \begin{tabularx}{\linewidth}{@{}l l l c c c *{4}{Y}@{}}
        \toprule
        \textbf{Tokenizer} & \textbf{Setting} & \textbf{Interp.} & \textbf{Params} & \textbf{Codebook} & \textbf{Dim} & \textbf{rFID}$\downarrow$ & \textbf{PSNR}$\uparrow$ & \textbf{SSIM}$\uparrow$ & \textbf{L1}$\downarrow$ \\
        \midrule
        \multirow{5}{*}{VQ-16~\cite{llamagen}} & Reported & -- & \multirow{5}{*}{71.9M} & \multirow{5}{*}{16384} & \multirow{5}{*}{8} & 2.19 & 20.79 & \textcolor{gray}{0.67$^{*}$} & -- \\
         & Reproduced & bilinear & & & & 2.10 & 21.51 & 0.59 & 0.0585 \\
         & Reproduced & bicubic  & & & & 2.19 & 20.79 & 0.55 & 0.0633 \\
         & Warm-up    & bicubic  & & & & 1.72 & 21.06 & 0.57 & -- \\
         & GEAR       & bicubic  & & & & 1.64 & 20.78 & 0.56 & -- \\
        \midrule
        \multirow{5}{*}{LFQ-16~\cite{openmagvitv2}} & Reported & bilinear & \multirow{5}{*}{115.1M} & \multirow{5}{*}{16384} & \multirow{5}{*}{14} & 2.55 & 22.21 & 0.62 & -- \\
         & Reproduced & bilinear & & & & 2.56 & 22.25 & 0.61 & 0.0524 \\
         & Reproduced & bicubic  & & & & 2.82 & 21.47 & 0.58 & 0.0571 \\
         & Warm-up    & bicubic  & & & & 2.42 & 20.97 & 0.56 & -- \\
         & GEAR       & bicubic  & & & & 2.13 & 20.48 & 0.55 & -- \\
        \midrule
        \multirow{5}{*}{IBQ-16~\cite{ibq}} & Reported & bilinear & \multirow{5}{*}{111.0M} & \multirow{5}{*}{16384} & \multirow{5}{*}{256} & 2.06 & 22.01 & 0.61 & -- \\
         & Reproduced & bilinear & & & & 2.05 & 22.04 & 0.61 & 0.0542 \\
         & Reproduced & bicubic  & & & & 2.23 & 21.23 & 0.58 & 0.0593 \\
         & Warm-up    & bicubic  & & & & 1.97 & 21.18 & 0.58 & -- \\
         & GEAR       & bicubic  & & & & 1.72 & 20.92 & 0.57 & -- \\
        \bottomrule
    \end{tabularx}
\end{table}

\section{Classifier-Free Guidance Sweep for Text-to-Image}
\label{app:t2i_cfg}

Under the controlled GPIC setting (\cref{tab:gpic_t2i_performance,tab:t2i_benchmarks}) we sweep the classifier-free guidance (CFG) scale from $1$ to $20$ on DPG-Bench and on GenEval (short and long prompts), for both the non-end-to-end tokenizer (LlamaGen-REPA) and the end-to-end one (GEAR). \cref{tab:t2i_cfg_sweep} reports the full sweep.

\begin{table}[t]
    \centering
    \setlength{\tabcolsep}{3pt}
    \caption{\textbf{Classifier-free guidance sweep on the GPIC-trained text-to-image models.} Each block reports one benchmark for the \emph{non-e2e} tokenizer (LlamaGen-REPA) and the \emph{e2e} tokenizer (GEAR) across CFG scales $1$--$20$. GenEval is shown for its original (short) and LLM-refined (long) prompts. The best of each row is in \textbf{bold} and the runner-up is \underline{underlined}.}
    \label{tab:t2i_cfg_sweep}
    \resizebox{\linewidth}{!}{%
    \begin{tabular}{ll|cccccccccccc}
    \toprule
    \multirow{2}{*}{\textbf{Benchmark}} & \multirow{2}{*}{\textbf{Tokenizer}} & \multicolumn{12}{c}{\textbf{Classifier-free guidance scale}} \\
    \cmidrule(lr){3-14}
     & & 1.0 & 1.75 & 2.0 & 4.0 & 6.0 & 8.0 & 10.0 & 12.0 & 14.0 & 16.0 & 18.0 & 20.0 \\
    \midrule
    \multirow{2}{*}{DPG-Bench} & non-e2e & 55.386 & 64.524 & 64.928 & 68.357 & 69.927 & 70.118 & 70.634 & 70.336 & 71.017 & 70.363 & \textbf{71.229} & \underline{71.062} \\
     & e2e & 55.369 & 63.782 & 64.815 & 69.199 & 71.281 & 71.511 & 72.134 & 72.383 & \underline{72.857} & \textbf{72.881} & 72.753 & 72.806 \\
    \midrule
    \multirow{2}{*}{GenEval (short)} & non-e2e & 0.0743 & 0.1400 & 0.1661 & 0.2481 & 0.2443 & 0.2572 & 0.2617 & 0.2739 & 0.2708 & 0.2717 & \underline{0.2814} & \textbf{0.2832} \\
     & e2e & 0.0861 & 0.1627 & 0.1862 & 0.2722 & 0.2964 & 0.3184 & 0.3149 & 0.3269 & 0.3245 & \underline{0.3339} & 0.3295 & \textbf{0.3413} \\
    \midrule
    \multirow{2}{*}{GenEval (long)} & non-e2e & 0.2181 & 0.3117 & 0.3284 & 0.3817 & 0.3854 & 0.4028 & 0.4164 & 0.4096 & 0.4163 & \textbf{0.4187} & 0.4101 & \underline{0.4182} \\
     & e2e & 0.2269 & 0.3401 & 0.3605 & 0.4228 & 0.4431 & 0.4595 & 0.4579 & \underline{0.4767} & 0.4601 & \textbf{0.4779} & 0.4701 & 0.4705 \\
    \bottomrule
    \end{tabular}%
    }
\end{table}

\myparagraph{Observations.}
The end-to-end tokenizer (GEAR) is the stronger of the two across essentially the entire sweep. On DPG-Bench the two are within noise at very low guidance ($\text{CFG}\le2$), but from $\text{CFG}\ge4$ GEAR pulls ahead by a stable $+1.3$ to $+2.5$ points. Its best score is $72.881$ at $\text{CFG}{=}16$, against $71.229$ at $\text{CFG}{=}18$ for the non-e2e tokenizer ($+1.65$ best-vs-best), reaching the higher peak at a \emph{smaller} guidance scale. On GenEval the advantage holds at every scale and for both prompt sets, with a roughly constant gap of $\approx{+}0.05$ (short) and $\approx{+}0.06$ (long). The long-prompt scores peak near $\text{CFG}{=}16$, whereas the short-prompt scores are still climbing at $\text{CFG}{=}20$.

\myparagraph{Discussion.}
A salient feature of these curves is that quality keeps rising up to very large guidance scales. A well-fit text-to-image model rarely needs a CFG of $16$--$20$, so the fact that performance is still improving there indicates that the single-epoch GPIC model is \emph{underfitting} and does not yet exploit the text condition strongly. This is expected given the deliberately small training budget. We stress that the goal of this study is a strictly controlled comparison that isolates the effect of the (end-to-end) tokenizer, not to chase state of the art: under this matched recipe GEAR's end-to-end tokenizer is consistently better across the whole guidance range and reaches higher peaks at smaller CFG, mirroring the class-conditional results. We expect this advantage to carry over once the recipe is scaled with longer training and stronger text conditioning, which we leave to future work.

\section{Per-Layer Representation Analysis}
\label{app:repr_peaks}

In \cref{fig:repr_analysis}, the image-level similarities (a,b) and Moran's~$I$ (c) peak near the alignment depth ($\ell{=}8$), whereas the patch-level curves (d,e) peak only in deeper layers. This is expected rather than anomalous. Image-level similarity reflects the global semantics that REPA injects directly at layer~$8$. Patch-level similarity instead reflects intra-image spatial structure, which only takes shape once deeper, causal layers have accumulated enough spatial context. We also probe the raw, pre-projection features, so the alignment layer need not be the most DINO-like in raw space. The bump in the FFT high-frequency ratio (f) at $\ell{=}8$ has the same origin: REPA constrains only the \emph{projected} feature $g_\phi(\mathbf{h}^{(8)})$ by cosine direction, so the raw alignment-layer feature is free to carry the injected per-token semantics in a high-frequency form, which the deeper causal layers then smooth into the spatially coherent, DINOv2-like structure that actually drives generation. Importantly, at \emph{every} depth GEAR dominates LlamaGen-REPA at the patch level and in both locality statistics (including this bump), which is what matters for generation.

\end{document}